\documentclass[runningheads]{llncs}

\usepackage[mobile]{eccv}

\usepackage{color}
\usepackage{colortbl}
\usepackage{afterpage}

\usepackage{eccvabbrv}

\usepackage{graphicx}
\usepackage{booktabs}

\usepackage[accsupp]{axessibility}  

\usepackage{hyperref}

\usepackage{orcidlink}

\usepackage{marvosym}

\newcommand{\bB}{\mathbf{B}}
\newcommand{\bc}{\mathbf{c}}
\newcommand{\bd}{\mathbf{d}}
\newcommand{\bE}{\mathbf{E}}
\newcommand{\bF}{\mathbf{F}} 

\newcommand{\bI}{\mathbf{I}}

\newcommand{\bP}{\mathbf{P}}

\newcommand{\bx}{\mathbf{x}}

\newcommand{\bOmega}{\boldsymbol{\Omega}}

\newcommand{\nR}{\mathbb{R}}
\newcommand{\nS}{\mathbb{S}}

\newcommand{\cL}{\mathcal{L}}

\newcommand{\figref}[1]{Fig.~\ref{#1}}

\newcommand{\eqnref}[1]{Eq.~\eqref{#1}}
\newcommand{\tabnref}[1]{Table~\ref{#1}}

\makeatletter
\DeclareRobustCommand\onedot{\futurelet\@let@token\@onedot}
\def\@onedot{\ifx\@let@token.\else.\null\fi\xspace}
 
\def\ie{i.e\onedot}

\makeatother

\newcommand{\PAR}[1]{\vspace{0.1cm}\noindent{\bf #1} }

\newcommand{\norm}[1]{\left\lVert#1\right\rVert}

\begin{document}

\title{BeNeRF: Neural Radiance Fields from a Single Blurry Image and Event Stream} 

\titlerunning{BeNeRF}

\author{Wenpu Li\inst{1,5}\textsuperscript{*}\orcidlink{0009-0009-6794-1810} \and
Pian Wan\inst{1,2}\textsuperscript{*}\orcidlink{0009-0007-9368-3662} \and
Peng Wang\inst{1,3}\textsuperscript{*}\orcidlink{0009-0003-5747-3319} \and
Jinghang Li\inst{4}\orcidlink{0000-0001-6196-6165} \and
Yi Zhou \inst{4}\orcidlink{0000-0003-3201-8873} \and
Peidong Liu\inst{1}\textsuperscript{$\dag$}\orcidlink{0000-0002-9767-6220}
}

\authorrunning{W. Li, P. Wan, P. Wang et al.}

\institute{\textsuperscript{1}Westlake University, \quad \textsuperscript{2}EPFL, \quad \textsuperscript{3}Zhejiang University, \quad \textsuperscript{4}Hunan University \textsuperscript{5}Guangdong University of Technology \\ \ \\
\href{https://github.com/wu-cvgl/BeNeRF}{https://github.com/wu-cvgl/BeNeRF}
}


\maketitle

\renewcommand{\thefootnote}{\fnsymbol{footnote}}
\footnotetext[1]{Equal contribution: \{liwenpu, wangpeng\}@westlake.edu.cn, pianwan@outlook.com}
\footnotetext[4]{Corresponding author: Peidong Liu(liupeidong@westlake.edu.cn).}


\begin{center}
    \setlength\tabcolsep{1pt}
    \includegraphics[width=0.93\linewidth]{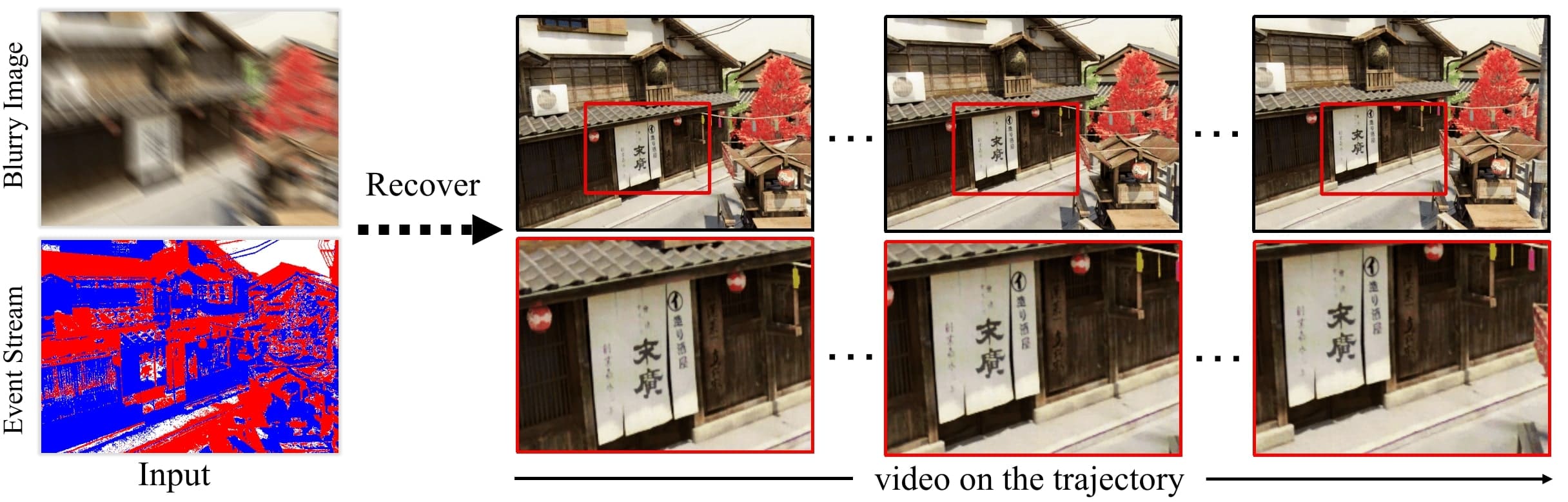}
    \captionof{figure}{Given a single blurry image and its corresponding event stream, BeNeRF can synthesize high-quality novel images along the camera trajectory, recovering a sharp and coherent video from the single blurry image.}
    \label{fig_teaser}
\end{center}

\begin{abstract}
Implicit scene representation has attracted a lot of attention in recent research of computer vision and graphics. Most prior methods focus on how to reconstruct 3D scene representation from a set of images.  
%
In this work, we demonstrate the possibility to recover the neural radiance fields (NeRF) from a single blurry image and its corresponding event stream. To eliminate motion blur, we introduce event stream to regularize the learning process of NeRF by accumulating it into an image. We model the camera motion with a cubic B-Spline in SE(3) space. Both the blurry image and the brightness change within a time interval, can then be synthesized from the NeRF given the 6-DoF poses interpolated from the cubic B-Spline. Our method can jointly learn both the implicit scene representation and the camera motion by minimizing the differences between the synthesized data and the real measurements without any prior knowledge of camera poses.
%
We evaluate the proposed method with both synthetic and real datasets. The experimental results demonstrate that we are able to render view-consistent latent sharp images from the learned NeRF and bring a blurry image alive in high quality. 


\end{abstract}
\section{Introduction}
\label{sec:intro}

Neural Radiance Fields (NeRF) \cite{mildenhall2020nerf} has drawn much attention due to its extraordinary ability in representing 3D scenes and synthesizing novel views. Given multi-view sharp RGB images and calibrated camera poses from COLMAP \cite{schoenberger2016sfm}, NeRF takes corresponding 3D spatial location and 2D view direction as input, and optimizes a multi-layer perception (MLP) to represent the 3D scene. More recent advanced methods also exploit explicit octree \cite{yu2021plenoctrees}, multi-resolution hash encoding \cite{muller2022instant} etc., to represent the 3D scene to improve both the training and rendering efficiency.

Prior methods usually rely on multi-view images to learn the 3D representation. Several pioneering works recently attempt to exploit a single image to learn the underlying neural radiance fields \cite{yu2021pixelnerf, rematas2021sharf, cai2022pix2nerf, rebain2022lolnerf}. They usually rely on a large dataset to pre-train the networks to learn priors to address the ill-posed problem. A blurry image further aggravates the problem due to the image quality degradation. Although motion blur is usually not preferred by most vision algorithms, they actually encode additional camera motion trajectory and more structural information compared to a sharp image. In this paper, we explore the possibility of recovering the neural radiance fields and camera motion trajectory from a single blurry image. Instead of learning priors from a large dataset as in previous works, we exploit the usage of an additional event stream to better constrain the problem.

Event stream can be acquired by an event camera \cite{lichtsteiner2008128} which captures pixel intensity changes caused by the relative motion between the static scene and camera. Unlike standard frame-based cameras, event camera captures asynchronous events with very low latency, leading to extremely high temporal resolution\cite{gallego2022eventbased}. This characteristic compensates with the image formation process of a blurry image (i.e. integral of photon measurements across time). Several prior works thus take advantage of both modalities for high quality single image deblurring \cite{pan2019bringing, wang2020event, sun2022eventbaseda}. However, these methods are unable to recover the camera motion trajectory and extract structural details from a single blurry image, thereby limiting their applicability in 3D computer vision tasks.
%
Some NeRF-based methods that incorporate event stream \cite{qi2023e2nerf, low2023_robust-e-nerf, klenk2022nerf, rudnev2023eventnerf, hwang2023ev} demonstrate the capability to achieve image deblurring and accurate reconstruction of neural radiance fields. Nonetheless, these methods necessitate input images from multiple viewpoints alongside event data.  
In contrast, we explore the usage of only a single blurry image for the NeRF recovery with unknown camera motions in this work. 



We represent the continuous camera motion with a cubic B-Spline in SE(3) space and define it as the trajectory of both frame-based camera and event camera. Given the neural 3D representation and interpolated poses from the cubic B-Spline, we can synthesize both the blurry image and the brightness change within a time interval via the physical image formation process. The NeRF and motion trajectory can then be jointly optimized by minimizing the difference between the synthesized data and the real measurements. 
To evaluate the performance of our method, we conduct experiments with both synthetic and real datasets. The experimental results demonstrate that our method is able to recover the neural radiance fields from a blurry image and its corresponding event stream without knowing prior knowledge of poses. We are thus able to render view-consistent latent sharp images encoded in a single blurred image from learned NeRF, effectively enhancing the quality of the blurry image. The experimental results further demonstrate that our method is even able to reach same performance as E$^2$NeRF \cite{qi2023e2nerf}, which targets for the same problem, but with multi-view images and longer event data. 
In summary, our key contributions are as follows:
\begin{itemize}
	\itemsep0em
	\item We propose a NeRF-based method that can recover the neural 3D representation from a single blurry image and its corresponding event stream, without knowing any ground truth poses;
	\item Our method is able to estimate the complex continuous trajectory of camera motion during the imaging process from a single blurry image, providing multi-view geometric information;
	\item We experimentally validate that our approach is able to recover high quality latent sharp images and high frame-rate video from a single blurry image, without any generalization issue. Furthermore, we are able to reach same level of performance as E$^2$NeRF \cite{qi2023e2nerf}, which requires multi-view images and longer event data.
\end{itemize}

\section{Related Work}
\label{sec:related}

We roughly categorize our related works into three main areas: neural implicit scene representation, single image deblurring and event-enhanced image deblurring. 

\PAR{Neural Implicit Scene Representation.} 
NeRF has attracted lots of attention due to its powerful ability of implicit 3D scene representation and novel view synthesis \cite{mildenhall2020nerf}. Many following works have been proposed to improve NeRF's performance or extend NeRF to other fields.\cite{wang2022nerf, lin2021barf, bian2022nopenerf} jointly trained NeRF with inaccurate camera poses. \cite{ma2022deblurnerf, yu2021pixelnerf, deng2022depth, mildenhall2022rawnerf, niemeyer2022regnerf, li2023usbnerf} improved the performance of NeRF with degraded images, including noisy or few images etc. Recently, several event based NeRF \cite{rudnev2023eventnerf, klenk2023enerf, qi2023e2nerf, hwang2023ev} have also been proposed.

We will mainly focus on the methods that are most related to ours. BAD-NeRF \cite{wang2023badnerf} aims to recover true underlying 3D scene representation from multi-view blurry inputs and inaccurate camera poses estimated from COLMAP \cite{schoenberger2016sfm}. However, BAD-NeRF and its variants \cite{lee2023exblurf, zhao2024badgaussians} struggle to address situations where the input is limited to a single image, primarily due to the severe illness of the problem. 
Event-Enhanced NeRF (E$^2$NeRF) \cite{qi2023e2nerf} aims to recover the 3D scene representation from multiple blurry images and event streams. To train NeRF without optimizing the camera poses, E$^2$NeRF manually segments the event stream using preset parameters, aiming to recover sharp images individually through EDI \cite{pan2019bringing}. However, in cases of serve camera motion, the recovered images from event stream may still be blurry, leading to inaccurate pose estimation in COLMAP \cite{schoenberger2016sfm}. This two-stage approach introduces errors (\ie either from EDI \cite{pan2019bringing} or COLMAP \cite{schoenberger2016sfm}) and fails to accurately model the continuous camera trajectory. In contrast, we are able to train both the NeRF and the camera motion trajectory jointly from a single blurry image and its corresponding event stream, which is harder than prior methods.

\PAR{Single Image Deblurring.}
A blurred image can be formulated as the convolution result of a sharp image and a kernel. Therefore, classical approaches \cite{levin2009understanding, jia2007single, joshi2009image, shan2008high, xu2010two} generally regard the deblurring problem as a joint optimization of the blur kernel and the latent virtual sharp images.
With the development of deep learning, many learning based end-to-end debluring method have been proposed \cite{tao2018scalerecurrent, kupyn2019deblurganv2a, zamir2022restormera, nah2017deep, kupyn2018deblurgan, sun2015learning, jin2018learning}. These methods usually demonstrate better qualitative and quantitative deblurring results. However, learning based deblurring methods are usually trained on a large dataset which contains paired blurry-sharp images. They would thus usually have limited generalization performance to domain-shifted images. Since NeRF is a test-time optimization approach, our method does not have the generalization performance issue. 

\PAR{Event Enhanced Image Deblurring.}
Since event camera is able to capture high dynamic temporal information\cite{gallego2022eventbased}, prior methods usually exploit event measurements to enhance the image deblurring performance \cite{pan2019bringing, sun2022eventbaseda, wang2020event, sun2023event, xu2021motion, jiang2020learning, zhang2023generalizing}. 
EDI \cite{pan2019bringing} is a simple and effective model, which is able to generate a sharp video under various types of blur by solving a single variable non-convex optimization problem. Different from EDI \cite{pan2019bringing}, \cite{sun2022eventbaseda, wang2020event, sun2023event} design end-to-end neural networks for event enhanced image deblurring/frame interpolation. The critical distinction lies in our method's capacity to extract potential camera motion trajectories from the event stream, thereby enhancing subsequent 3D vision tasks with additional geometric insights.

\section{Methodology}
\label{sec:method}
\begin{figure*}[t]
	\centering
	\includegraphics[width=1\linewidth]{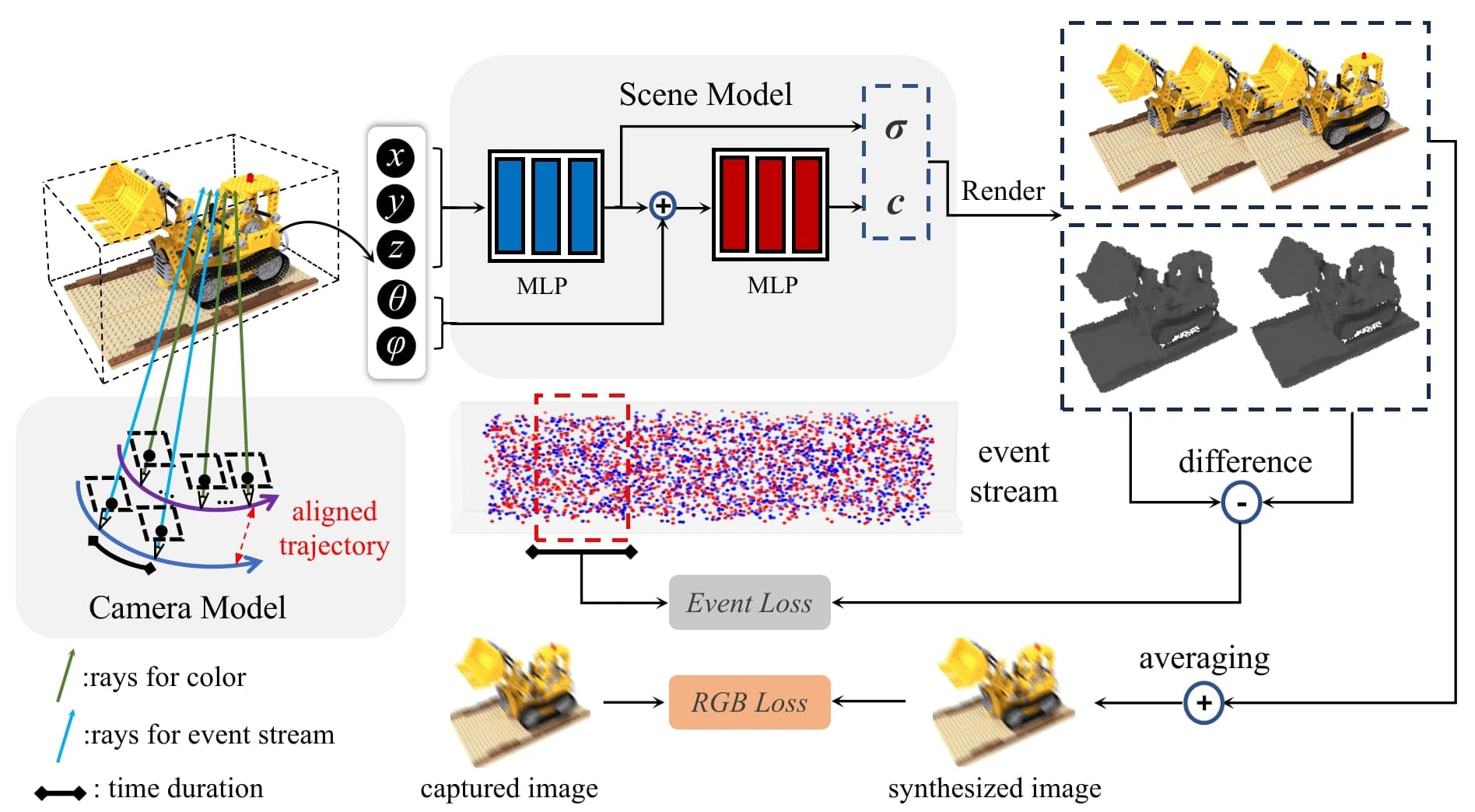}
    \caption{{\bf{The pipeline of our method.}} Given a motion blurry image and its corresponding event stream, we aim to recover both the implicit sharp scene representation and its camera motion trajectory within exposure time. We exploit a continuous time representation for the motion trajectory, and maximize the coherence between both the real measurements and synthesized data for the recovery.}
	\label{fig_method}
\end{figure*}

Given a single blurry image and its corresponding event stream, our method recovers the underlying 3D scene representation and the camera motion trajectory jointly. The details of our method are shown in \figref{fig_method}. In particular, we represent the 3D scene with neural radiance fields and the camera motion trajectory with a cubic B-Spline in SE(3) space. Both the blurry image and accumulated events within a time interval can thus be synthesized from the 3D scene representation providing the camera poses. The camera trajectory, NeRF, are then optimized by minimizing the difference between the synthesized data and the real measurements. The details are as follows. 

\subsection{Neural Implicit Representation}
\label{subsec:neural}
We adopt Multi-layer Perceptron (MLP) to represent the 3D scene as the original NeRF \cite{mildenhall2020nerf}. More advanced representations, such as multi-resolution hash encoding \cite{muller2022instant}, can also be exploited to further improve its performance. In particular, the scene model is represented by a learnable mapping function $\bF_{\theta }:(\bx, \bd) \rightarrow ( \mathbf{c}, \sigma)$, which requires a Cartesian coordinates $\bx \in \nR^3$ and a viewing direction $\bd \in \nS^2$ as input, and outputs the corresponding volume density $\sigma \in \nR$ and color $\bc \in \nR^3$. Both the 3D point $\bx$ and viewing direction $\bd$ are defined in the world coordinate frame. They are a function of the pixel coordinate, camera pose and the corresponding intrinsic parameters. To query the intensity of a pixel, we can apply volume rendering by sampling 3D points along the ray, which originates from the camera center and passes through the pixel. The volume rendering can be formally defined as follows:
\begin{equation}
	\bI(\bx) = \sum_{i=1}^{n} T_i(1-{\rm exp}(-\sigma_i\delta_i))\boldsymbol{c}_i,
\end{equation}
where $n$ is the number of sampled points along the ray, $\boldsymbol{c}_i$ and $\sigma_i$ are the predicted color and volume density of the $i^{th}$ sampled 3D point via $\bF_\theta$, $\delta_i$ is the distance between the $i^{th}$ and $(i+1)^{th}$ sampled point, $T_i$ is the transmittance which represents the probability that the ray does not hit any particle until the $i^{th}$ sampled point. $T_i$ can be computed via:
\begin{equation}
    T_i = {\textrm{exp}} (-\sum_{k=1}^{i-1}\sigma_k\delta_k).
\end{equation}

\subsection{Camera Motion Trajectory Modeling}
\label{subsec:camera_trajectroy}
We use a differentiable cubic B-Spline in SE(3) space to better model the camera motion trajectory. The spline is represented by a set of learnable control knots $\boldsymbol{T}_{c_i}^w \in \mathbb{SE}(3)$ for $i=0,1,...,n$. $\boldsymbol{T}_{c_i}^w$ represents the $i^{th}$ control knot, which is defined as a transformation matrix from the camera coordinate frame to world frame.  
For brevity, we denote $\boldsymbol{T}_{c_i}^w$ with $\boldsymbol{T}_i$ for subsequent derivations. We assume the control knots are sampled with a uniform time interval $\Delta t$ and the trajectory starts from $t_0$. Spline with a smaller $\Delta t$ can represent a smoother motion, with an expense of more control knots to optimize. Since four consecutive control knots determine the value of the spline curve at a particular timestamp, we can thus compute the starting index of the four control knots for time $t$ by:
\begin{equation}
	k = \lfloor \frac{t - t_0}{\Delta t} \rfloor,
\end{equation}
where $\lfloor * \rfloor$ is the floor operator. Then we can obtain the four control knots responsible for time $t$ as $\boldsymbol{T}_k$, $\boldsymbol{T}_{k+1}$, $\boldsymbol{T}_{k+2}$ and $\boldsymbol{T}_{k+3}$. We can further define $u = \frac{t - t_0}{\Delta t} - k$, where $u \in [0, 1)$ to transform $t$ into a uniform time representation. Using this time representation and based on the matrix representation for the De Boor-Cox formula \cite{qin1998general}, we can write the matrix representation of a cumulative basis $\mathcal{B}$(u) as 
\begin{equation}
	\mathcal{B}(u) = \mathcal{M} \begin{bmatrix}
		1 \\ u \\ u^2 \\ u^3
	\end{bmatrix},  \quad
    \mathcal{M} = \frac{1}{6} \begin{bmatrix}
    	6 & \ \ \ 0 & \ \ 0 & \ \ 0 \\
    	5 & \ \ \ 3 & \ -3 &  \ \ 1 \\
    	1 & \ \ \ 3 & \ \ 3 & \ -2 \\
    	0 & \ \ \ 0 & \ \ 0 & \ \ 1
    \end{bmatrix}.
\end{equation}
The pose at time $t$ can be computed as:
\begin{equation}\label{eq_rgb}
	\boldsymbol{T}(u) = \boldsymbol{T}_k \cdot \prod_{j=0}^2 \mathrm{exp}(\mathcal{B}(u)_{j+1} \cdot \bOmega_{k+j}),
\end{equation}
where $\mathcal{B}(u)_{j+1}$ denotes the $(j+1)^{th}$ element of the vector $\mathcal{B}(u)$, $\bOmega_{k+j} = \mathrm{log}(\boldsymbol{T}_{k+j}^{-1} \cdot \boldsymbol{T}_{k+j+1}))$.

Since we only consider the continuous camera motion corresponding to a single blurry image, the time interval is thus usually short. We found that four control knots would be sufficient to deliver satisfying results. Therefore, we use the minimal configuration for the following experiments and they are initialized randomly around the identity pose. 

\subsection{Blurry Image Formation Model}
A motion blurred image $\bB(\bx) \in \nR^{\mathrm{W} \times \mathrm{H} \times 3}$ is physically formed by collecting photons during the exposure time and it can be mathematically modeled as:
\begin{equation} \label{eq_blurry}
	\bB(\bx)  \approx \frac{1}{n} \sum_{i=0}^{n-1} \bI_\mathrm{i}(\bx), 
\end{equation}
where both $\mathrm{W}$ and $\mathrm{H}$ are the width and height of the image respectively, $n$ is the number of sampled images, $\bx \in \nR^2$ represents the pixel location, $\bI_\mathrm{i}(\bx) \in \nR^{\mathrm{W} \times \mathrm{H} \times 3}$ is the $i^{th}$ virtual sharp image sampled within the exposure time. The virtual sharp images can be rendered from the neural 3D scene representation along the previous defined camera trajectory. It can be seen that $\bB(\bx)$ is differentiable with respect to the parameters of NeRF and the motion trajectory. 

\subsection{Event Data Formation Model}
An event camera records changes of the brightness as a stream of events asynchronously. Every time a pixel brightness change reaches a contrast threshold (\ie $|L(\bx, t_i + \delta t) - L(\bx, t_i)| \ge C$), the camera will trigger an event $e_i=(\bx,t_i,p_i)$, where $p_i \in (-1, +1)$ is the polarity of the event, $L(\bx, t_i) = \log(\bI(\bx, t_i))$ is the brightness logarithm of pixel $\bx$ at timestamp $t_i$, $C$ is the contrast threshold. 

To relate NeRF representation with the event stream, we accumulate the real measured events within a time interval $\Delta t$ to an image $\bE(\bx)$. The accumulation is defined as:
\begin{equation}\label{eq_event_accumulate}
	\bE(\bx) = C \{e_i(\bx,t_i,p_i)\}_{t_k<t_i<t_k+\Delta t},
\end{equation}
where $e(\bx, t_i, p_i)$ is the $i^{th}$ event within the defined time interval corresponding to pixel $\bx$. For real event cameras, the contrast threshold $C$ changes over time and varies pixel by pixel. We therefore normalize the accumulated event as in \cite{hidalgo-carrio2022eventaided, klenk2023enerf} to eliminate the effect of unknown $C$:
\begin{equation}\label{eq_event_normalize}
	\bE_n(\bx) = \frac{\bE(\bx)}{\norm{\bE(\bx)}_2},
\end{equation}
Given the interpolated start pose and end pose corresponding to the time interval $\Delta t$ from the spline, we are able to render two gray-scale images (\ie $\bI_{start}$ and $\bI_{end}$) from NeRF. The synthesized accumulated event image $\hat{\bE}$ can then be computed as:
\begin{equation}
	\hat{\bE}(\bx) = \log(\bI_{end}(\bx)) - \log(\bI_{start}(\bx)),
\end{equation}
where $\hat{\bE}(\bx)$ depends on the parameters of both the cubic spline and NeRF, and is differentiable with respect to them. We can also normalize $\hat{\bE}(\bx)$ to $\hat{\bE}_n(\bx)$ similarly as in \eqnref{eq_event_normalize} for loss computation.  

\subsection{Loss Functions}
We minimize the sum of a photo-metric loss $\cL_{p}$ and an event loss $\cL_{e}$:
\begin{equation}
    \cL_{total} = \cL_{p} + \beta\cL_{e},
\end{equation}
where $\cL_{p}$ represents the loss for the frame-based camera, $\cL_{e}$ represents the loss for the accumulated events within a randomly sampled time interval, and $\beta$ is a hyper-parameter. Both losses are defined as follows:
\begin{align}
	&\cL_{p} = \norm{\bB(\bx) - \hat{\bB}(\bx)}^2, \\
	&\cL_{e} = \norm{ \bE_n(\bx) - \hat{\bE}_n(\bx) }^2,
\end{align}
where $\hat{\bB}(\bx)$ is the real captured blurry image.

\section{Experiments} \label{sec:exp}

\subsection{Experimental Setup}
\PAR{Synthetic datasets.} We generate synthetic datasets for both quantitative and qualitative evaluations via Unreal Engine \cite{unreal} and Blender\cite{foundationblender}. To have a more realistic synthesis, we interpolate the real camera motion trajectories from ETH3D \cite{schops2019bad} to render high frame-rate images. In total, we generate three sequences (\ie livingroom, whiteroom and pinkcastle) via Unreal Engine and two sequences (\ie tanabata and outdoorpool) via Blender. For thorough evaluations, we synthesized twenty blurry images and corresponding event streams for each sequence. The event streams are generated via ESIM \cite{rebecq2018esim} from high frame-rate video. Furthermore, we additionally employed the synthetic dataset proposed by E$^2$NeRF to compare our method with NeRF-based image-deblur methods which require multi-view training data. This dataset includes synthesized blurry images paired with their respective event streams, which expands upon the six scenes (\ie chair, ficus, hotdog, lego, materials and mic).

\PAR{Real-world datasets.} We utilized the real-world datasets proposed by E$^2$NeRF, captured using the DAVIS346 color event camera in real-world scenarios. The dataset encompass five challenging scenes (\ie letter, lego, camera, plant and toys). The exposure time for RGB frames was set to 100ms, resulting in occurrences of complex camera motion and severe motion blur within the time interval.

\PAR{Baseline methods and evaluation metrics.} To evaluate the performance of our method in terms of image deblurring, we compare it against state-of-the-art deep learning-based single image deblur methods, i.e. SRN-Deblur\cite{tao2018scalerecurrent}, HINet\cite{chen2021hinet}, DeblurGANv2\cite{kupyn2019deblurganv2a}, MPRNet\cite{zamir2021multistage}, NAFNet\cite{chen2022simple} and Restormer\cite{zamir2022restormera}, as well as event-enhanced single image-based deblur methods, i.e. EDI\cite{pan2019bringing}, eSLNet \cite{wang2020event}. We also compared our method with NeRF-based image deblur method using multi-view information. The quality of the rendered image is evaluated with the commonly used PSNR, SSIM and LPIPS \cite{zhang2018unreasonable} metrics. Since the lack of sharp reference images in real-world datasets, we conducted quantitative analysis experiments on five real scenes using the no-reference image quality assessment metrics BRISQUE \cite{mittal2012no}.

\PAR{Implementation details.} We implement our method with PyTorch. The implicit representation of the scene from MLP (i.e. $F_\theta$) is built from NeRF\cite{mildenhall2020nerf} without any modification. We randomly initialize trajectory control knots for cameras within a range of $(0, 0.01)$. We use two separate Adam optimizers \cite{kingma2017adama} for the scene model (i.e. $F_\theta$) and camera motion (i.e. $\bP_i$). The learning rate for the scene model and poses decay from $5 \times 10^{-4}$ with a rate of $0.1$ exponentially. In each training step, 1024 pixels for brightness and 1024 pixels for color are sampled. The weight of the event loss $\beta$ is selected to be $0.1$ for synthetic datasets and $2$ for real-world datasets, respectively. We train our model for 80K iterations for each image and its corresponding event stream.

\begin{table}[t]
    \caption{{\bf{Ablation studies on the number of virtual sharp images.}} The results demonstrate that the image quality gradually saturates as the number of virtual sharp images increases. By compromising the image quality and computational efficiency, we select $n=19$ for all our experiments.}
	\centering
	\setlength\tabcolsep{11.0pt}
    \resizebox{0.9\linewidth}{!}{
		\begin{tabular}{c|ccc|ccc}
			& \multicolumn{3}{c|}{Livingroom} & \multicolumn{3}{c}{Tanabata}\\
			\hline
			$n$ & PSNR$\uparrow$ & SSIM$\uparrow$ & LPIPS$\downarrow$ & PSNR$\uparrow$ & SSIM$\uparrow$ & LPIPS$\downarrow$ \\
			\specialrule{0.05em}{1pt}{1pt}
			7 & 36.16 & .9291 & .0711 & 29.28 & .8533 & .0712 \\
			11 & \cellcolor{yellow!25}36.84 & .9353 & .0659 & 30.77 & .8788 & .0622 \\
			15 & \cellcolor{orange!25}37.03 & \cellcolor{yellow!25}.9368 & \cellcolor{yellow!25}.0635 & \cellcolor{yellow!25}31.90 & \cellcolor{yellow!25}.8979 & \cellcolor{yellow!25}.0529 \\
   
			19 & \cellcolor{red!25}37.11 & \cellcolor{orange!25}.9370 & \cellcolor{orange!25}.0632 & \cellcolor{orange!25}32.14 & \cellcolor{orange!25}.9015 & \cellcolor{orange!25}.0515 \\
   
			23 & \cellcolor{red!25}37.11 & \cellcolor{red!25}.9375 & 
            \cellcolor{red!25}.0629 & \cellcolor{red!25}32.35 & 
            \cellcolor{red!25}.9042 & \cellcolor{red!25}.0506 \\
		\end{tabular}
	}
	\label{table_rgb_n}
\end{table}

\begin{table}[t]
	\centering
    \caption{{\bf{Ablation studies on event accumulation time lengths.}} The experimental results demonstrate that the image quality can be affected by the time length. For all experiments, we select $\alpha=0.1$ for event data accumulation.}
	\setlength\tabcolsep{11.0pt}
    \resizebox{0.9\linewidth}{!}{
		\begin{tabular}{c|ccc|ccc}
			& \multicolumn{3}{c|}{Livingroom} & \multicolumn{3}{c}{Tanabata}\\
			\hline
			$\alpha$ & PSNR$\uparrow$ & SSIM$\uparrow$ & LPIPS$\downarrow$ & PSNR$\uparrow$ & SSIM$\uparrow$ & LPIPS$\downarrow$ \\
			\specialrule{0.05em}{1pt}{1pt}
			0.05 & 36.66 & .9325 & .0724 & 31.86 & .8977 & .0538 \\
   
			0.10 & \cellcolor{yellow!25}37.11 & \cellcolor{orange!25}.9370 & 
                    .0632 & 32.14 & 
                    \cellcolor{yellow!25}.9015 & .0515 \\
                    
			0.15 & \cellcolor{red!25}37.21 & \cellcolor{red!25}.9376 & \cellcolor{yellow!25}.0601 & \cellcolor{yellow!25}32.19 & 
            \cellcolor{red!25}.9019 & \cellcolor{yellow!25}.0509 \\
            
			0.20 & \cellcolor{orange!25}37.16 & \cellcolor{yellow!25}.9369 & \cellcolor{orange!25}.0596 & \cellcolor{orange!25}32.20 & \cellcolor{orange!25}.9016 & \cellcolor{orange!25}.0499 \\
   
			0.25 & \cellcolor{yellow!25}37.11 & .9358 & 
            \cellcolor{red!25}.0589 & \cellcolor{red!25}32.21 & 
            \cellcolor{yellow!25}.9015 & \cellcolor{red!25}.0496 \\
		\end{tabular}
	}
	\label{table_event_alpha}
\end{table}

\begin{table}[t]
    \centering
    \caption{{\bf{Ablation studies on trajectory representations.}} The results demonstrate that cubic B-spline can deliver better performance than linear interpolation.}
    \setlength\tabcolsep{6.5pt}
    \resizebox{0.9\linewidth}{!}{
        \begin{tabular}{c|ccc|ccc}
            & \multicolumn{3}{c|}{Livingroom} & \multicolumn{3}{c}{Tanabata}\\
            & PSNR$\uparrow$ & SSIM$\uparrow$ & LPIPS$\downarrow$ & PSNR$\uparrow$ & SSIM$\uparrow$ & LPIPS$\downarrow$ \\
            \specialrule{0.05em}{1pt}{1pt}
            linear interpolation & 34.16 & .9035 & .1133 & 27.42 & .8164 & .1117 \\
            cubic B-Spline & \cellcolor{red!25}37.11 & \cellcolor{red!25}.9370 & \cellcolor{red!25}.0632 & \cellcolor{red!25}32.14 & \cellcolor{red!25}.9015 & \cellcolor{red!25}.0515 \\
    \end{tabular}
    }
    \label{table_trajectory}
\end{table}

\subsection{Ablation Study}
We evaluate the performance of our method under various configurations. To quantify the differences, we exploit two synthetic datasets (\ie Livingroom and Tanabata) for the experiments. 

\PAR{Effect of the number of virtual sharp images.}  We evaluate the effect of different numbers of the interpolated virtual images as mentioned in \eqnref{eq_blurry}. The experimental results are presented in \tabnref{table_rgb_n}. It demonstrates that more virtual images deliver better image quality, while requiring more computational resources. By compromising the image quality and computation requirement, we choose $n=19$ for our experiments.

\PAR{Effect of time lengths for event accumulation.} We study the effect of different time lengths $\Delta t$ for event accumulation as in \eqnref{eq_event_accumulate}. The timestamps of the event stream are normalized to a range of $(0, 1)$ by its total time length. We choose different values from 0.05 to 0.25 for the ablation study. The experimental results are presented in \tabnref{table_event_alpha}, showing that the performance on Tanabata dataset gradually saturates as $\alpha$ increases until to 0.25, whereas on Livingroom dataset, performance initially improves with increasing $\alpha$ but subsequently declines. This may depend on the noise level in event stream of different time lengths. In all experiments, we select $\alpha = 0.1$. 

\PAR{Effect of trajectory representations.} We explore the difference between linear interpolation and cubic B-Spline to represent the motion trajectory. The experimental results shown in \tabnref{table_trajectory} demonstrate that cubic B-Spline deliver better performance on complex motions than that of the linear interpolation. We exploit cubic B-Spline for the experiments. 

\begin{table}[t]
    \centering
    \caption{{\textbf{Quantitative comparisons on single image deblurring with synthetic datasets.}} The results demonstrate that our method performs better than prior learning-based methods in terms of image quality. For HINet and NAFNet, we tests pre-trained weights from both GoPro and REDS datasets(*). Due to page limits, the results of the SSIM metric can be found in the supplementary materials.}
    \resizebox{0.9\linewidth}{!}{
        \begin{tabular}{c|cccccc}
            & \multicolumn{6}{c}{PSNR $\uparrow$} \\
            & Livingroom & Whiteroom & Pinkcastle & Tanabata & Outdoorpool & \textit{Average} \\
            
            \specialrule{0.05em}{1pt}{1pt}
            
            DeblurGANv2 \cite{kupyn2019deblurganv2a} & $29.26$ & \cellcolor{yellow!25}$27.64$ & \cellcolor{orange!25}$23.16$ & \cellcolor{orange!25}$20.09$ & $26.89$ &\cellcolor{yellow!25}$25.41$ \\

            SRN-deblur \cite{tao2018scalerecurrent} &\cellcolor{orange!25}$30.86$ &$27.59$ &\cellcolor{yellow!25}$23.12$ & \cellcolor{yellow!25}$19.89$ & \cellcolor{orange!25}$27.79$ &\cellcolor{orange!25}$25.85$ \\

            MPRNet \cite{zamir2021multistage} &$28.57$ & $26.49$ & $21.60$ & $18.20$ & $27.02$ & $24.38$ \\

            HINet \cite{chen2021hinet} &$28.56$ & $26.27$ & $21.91$ & $18.59$ & $26.70$ & $24.41$ \\

            HINet* \cite{chen2021hinet} &$27.55$ & $22.89$ & $20.25$ & $18.15$ & $27.14$ & $23.20$ \\
            NAFNet \cite{chen2022simple} &\cellcolor{yellow!25}$29.92$ & \cellcolor{orange!25}$28.16$ & $22.41$ & $18.96$ & $26.75$ & $25.24$ \\
            NAFNet* \cite{chen2022simple} &$28.18$ & $23.67$ & $20.85$ & $18.38$ & \cellcolor{yellow!25}$27.52$ & $23.72$ \\
            Restormer \cite{zamir2022restormera} & $29.48$ & $27.39$ & $22.22$ & $18.82$ & $27.35$ & $25.05$ \\

            \specialrule{0.05em}{1pt}{1pt}
            
            \textbf{BeNeRF} & \cellcolor{red!25}$37.11$ & \cellcolor{red!25}$32.95$ & \cellcolor{red!25}$29.68$ & \cellcolor{red!25}$32.14$ & \cellcolor{red!25}$36.38$ & \cellcolor{red!25}$33.65$ \\
        
        \end{tabular}
    }

    \resizebox{0.9\linewidth}{!}{
        \begin{tabular}{c|cccccc}
            & \multicolumn{6}{c}{LPIPS $\downarrow$} \\
            & Livingroom & Whiteroom & Pinkcastle & Tanabata & Outdoorpool & \textit{Average} \\
            \specialrule{0.05em}{1pt}{1pt}
            
            DeblurGANv2 \cite{kupyn2019deblurganv2a} &\cellcolor{orange!25}.2087 &\cellcolor{orange!25}.1989 &\cellcolor{orange!25}.2608 &\cellcolor{yellow!25}.3934 &\cellcolor{orange!25}.3100 &\cellcolor{orange!25}.2744 \\

            SRN-deblur \cite{tao2018scalerecurrent} &.2529 &.2503 &.3245 & .4260 & .3594 & .3226 \\

            MPRNet \cite{zamir2021multistage} & .2621 & .2564 & .3586 & .4173 & .3679 & .3325 \\

            HINet \cite{chen2021hinet} & .2468 & .2620 & .3500 & .4024 & .3355 & .3193 \\

            HINet* \cite{chen2021hinet} & .3327 & .3602 & .3789 & .5265 & .4397 & .4076 \\
            NAFNet \cite{chen2022simple} & \cellcolor{yellow!25}.2268 &\cellcolor{yellow!25} .1991 & \cellcolor{yellow!25}.3058 & \cellcolor{orange!25}.3908 & \cellcolor{yellow!25}.3280 & \cellcolor{yellow!25}.2901 \\
            NAFNet* \cite{chen2022simple} & .3182 & .3566 & .3943 & .5271 & .4257 & .4044 \\
            Restormer \cite{zamir2022restormera} & .2391 & .2493 & .3373 & .4248 & .3664 & .3234 \\

            \specialrule{0.05em}{1pt}{1pt}
            
            \textbf{BeNeRF} &\cellcolor{red!25}.0632 &\cellcolor{red!25}.0788 &\cellcolor{red!25}.0761 &\cellcolor{red!25}.0515 &\cellcolor{red!25}.0677 &\cellcolor{red!25}.0675 \\
        
        \end{tabular}
    }
	\label{table_quant_sinimage}
\end{table}

\subsection{Quantitative evaluations}
To evaluate the performance of our method, we compare it against single-image deblurring methods, event-enhanced single-image deblurring methods, and NeRF-based image deblurring methods requiring multi-view information on both synthetic datasets and real datasets. The experimental results are presented in \tabnref{table_quant_sinimage}, \tabnref{table_quant_evsinimage}, \tabnref{table_quant_nerfbased} and \tabnref{table_quant_real}.

In particular, we compare against SRN-Deblur\cite{tao2018scalerecurrent}, DeblurGANv2\cite{kupyn2019deblurganv2a}, MPRNet\cite{zamir2021multistage}, HINet\cite{chen2021hinet}, NAFNet\cite{chen2022simple}, Restormer\cite{zamir2022restormera} in terms of single image deblurring. The experimental results shown in \tabnref{table_quant_sinimage} demonstrates that our method significantly outperforms prior state-of-the-art methods. It shows that prior learning-based methods have limited generalization performance on domain-shifted images, especially with large motion blurs. 

\begin{table}[t]
    \centering
    \setlength\tabcolsep{3pt}
    \caption{{\textbf{Quantitative comparisons on event-enhanced single image deblurring with synthetic datasets.}} The results demonstrate that our method performs better than both EDI and eSLNet. Due to page limits, the results of the SSIM metric can be found in the supplementary materials.}
    \resizebox{0.9\linewidth}{!}{
        \begin{tabular}{c|cccccc}
            & \multicolumn{6}{c}{PSNR $\uparrow$} \\
            & Livingroom & Whiteroom & Pinkcastle & Tanabata & Outdoorpool & \textit{Average} \\
            \specialrule{0.05em}{1pt}{1pt}
            
            eSLNet\cite{wang2020event} & \cellcolor{yellow!25}14.22 & \cellcolor{yellow!25}10.81 & \cellcolor{yellow!25}10.49 & \cellcolor{yellow!25}8.86 & \cellcolor{yellow!25}11.80 & \cellcolor{yellow!25}11.24 \\
            EDI\cite{pan2019bringing} & \cellcolor{orange!25}32.61 & \cellcolor{orange!25}30.33 & \cellcolor{orange!25}27.24 & \cellcolor{orange!25}24.87 & \cellcolor{orange!25}31.64 & \cellcolor{orange!25}29.34 \\
            
            \specialrule{0.05em}{1pt}{1pt}
            
            \textbf{BeNeRF} &\cellcolor{red!25}37.11 &\cellcolor{red!25}32.95 &\cellcolor{red!25}29.68 &\cellcolor{red!25}32.14 &\cellcolor{red!25}36.38 &\cellcolor{red!25}33.65 \\
        \end{tabular}
    }
    
    \resizebox{0.9\linewidth}{!}{
        \begin{tabular}{c|cccccc}
            & \multicolumn{6}{c}{LPIPS $\downarrow$} \\
            & Livingroom & Whiteroom & Pinkcastle & Tanabata & Outdoorpool & \textit{Average} \\
            \specialrule{0.05em}{1pt}{1pt}
            
            eSLNet\cite{wang2020event} & \cellcolor{yellow!25}.3981 & \cellcolor{yellow!25}.4236 & \cellcolor{yellow!25}.4902 & \cellcolor{yellow!25}.5067 & \cellcolor{yellow!25}.4676 & \cellcolor{yellow!25}.4572 \\
            EDI\cite{pan2019bringing} & \cellcolor{orange!25}.0904 & \cellcolor{orange!25}.1020 & \cellcolor{orange!25}.0779 & \cellcolor{orange!25}.1039 & \cellcolor{orange!25}.1409 & \cellcolor{orange!25}.1030 \\

            \specialrule{0.05em}{1pt}{1pt}
            
            \textbf{BeNeRF} &\cellcolor{red!25}.0632 &\cellcolor{red!25}.0788 &\cellcolor{red!25}.0761 &\cellcolor{red!25}.0515 &\cellcolor{red!25}.0677 &\cellcolor{red!25}.0675 \\
        
        \end{tabular}
    }
	\label{table_quant_evsinimage}
\end{table}

\begin{table}
    \centering
    \caption{{\textbf{Quantitative comparisons on NeRF-based image deblurring with synthetic datasets from E$^2$NeRF.}} The results indicate that our method outperforms both NeRF and Deblur-NeRF, and exhibits performance comparable to E$^2$NeRF in terms of the PSNR metric. Moreover, our method even surpasses E$^2$NeRF with the LPIPS metric. Due to page limits, the results of the SSIM metric can be found in the supplementary materials.}
    \setlength\tabcolsep{6pt}
    \resizebox{0.9\linewidth}{!}{
        \begin{tabular}{c|ccccccc}
            & \multicolumn{7}{c}{PSNR $\uparrow$} \\
            & Chair & Ficus & Hotdog & Lego & Materials & Mic & \textit{Average} \\
            \specialrule{0.05em}{1pt}{1pt}

            NeRF\cite{mildenhall2020nerf} & 24.29 & \cellcolor{yellow!25}22.98 & \cellcolor{yellow!25}27.75 & 21.95 & 19.99 & \cellcolor{yellow!25}20.50 & \cellcolor{yellow!25}22.91 \\
            Deblur-NeRF\cite{ma2022deblurnerf} & \cellcolor{yellow!25}25.87 & 22.86 & 24.62 & \cellcolor{yellow!25}24.47 & \cellcolor{yellow!25}20.54 & 11.92 & 21.71 \\
            E$^2$NeRF\cite{qi2023e2nerf} & \cellcolor{red!25}31.28 &\cellcolor{orange!25}30.00 & \cellcolor{red!25}34.34 & \cellcolor{red!25}28.11 & \cellcolor{orange!25}27.27 & \cellcolor{red!25}27.60 & \cellcolor{red!25}29.77 \\
            
            \specialrule{0.05em}{1pt}{1pt}
            
            \textbf{BeNeRF} &\cellcolor{orange!25}31.17 &\cellcolor{red!25}30.81 &\cellcolor{orange!25}34.31 &\cellcolor{orange!25}28.09 &\cellcolor{red!25}27.44 &\cellcolor{orange!25}26.13 & \cellcolor{orange!25}29.66\\
        
        \end{tabular}
    }
    
    \resizebox{0.9\linewidth}{!}{
        \begin{tabular}{c|ccccccc}
            & \multicolumn{7}{c}{LPIPS $\downarrow$} \\
            & Chair & Ficus & Hotdog & Lego & Materials & Mic & \textit{Average} \\
            \specialrule{0.05em}{1pt}{1pt}

            NeRF\cite{mildenhall2020nerf} & \cellcolor{yellow!25}.1254 & \cellcolor{yellow!25}.1037 & \cellcolor{yellow!25}.1158 & .2103 & \cellcolor{yellow!25}.1512 & \cellcolor{yellow!25}.1579 & \cellcolor{yellow!25}.1441 \\
            Deblur-NeRF\cite{ma2022deblurnerf} & .2185 & .1541 & .2138 & \cellcolor{yellow!25}.2053 & .2562 & .3706 & .2364 \\
            E$^2$NeRF\cite{qi2023e2nerf} & \cellcolor{orange!25}.0608 & \cellcolor{orange!25}.0362 & \cellcolor{orange!25}.0660 & \cellcolor{orange!25}.1078 & \cellcolor{orange!25}.0919 & \cellcolor{red!25}.0724 & \cellcolor{orange!25}.0725 \\
            
            \specialrule{0.05em}{1pt}{1pt}
            
            \textbf{BeNeRF} &\cellcolor{red!25}.0500 &\cellcolor{red!25} .0299 &\cellcolor{red!25}.0539 &\cellcolor{red!25}.0745 &\cellcolor{red!25}.0708 &\cellcolor{orange!25}.0738 & \cellcolor{red!25}.0588 \\
        
        \end{tabular}
    }
	\label{table_quant_nerfbased}
\end{table}

We also compare against prior event-enhanced single image deblurring methods, such as EDI\cite{pan2019bringing} and eSLNet\cite{wang2020event}. The results in \tabnref{table_quant_evsinimage} demonstrates that our method has superior performance when compared to them. eSLNet demonstrates poor generalization performance, since we are unable to fine-tune it on our evaluation datasets. It demonstrate the benefit on incorporating event streams to enhance single image deblurring task under the framework of NeRF. 

Furthermore, we conducted detailed comparisons with NeRF-based image deblurring methods that require multi-view information on the synthetic dataset proposed by E$^2$NeRF. We compared against NeRF\cite{mildenhall2020nerf}, Deblur-NeRF\cite{ma2022deblurnerf} and E$^2$NeRF \cite{qi2023e2nerf}. The experimental results in \tabnref{table_quant_nerfbased} demonstrate that despite utilizing only a single blurred image and event stream of a small time interval, our method achieves performance comparable to E$^2$NeRF \cite{qi2023e2nerf}, which utilizes multi-view images and a longer event stream, in terms of PSNR metric. Moreover, our method even surpasses E$^2$NeRF \cite{qi2023e2nerf} in terms of the LPIPS metric.

Finally, we select the best-performing algorithms on the synthetic dataset, excluding our method, from single-image deblurring methods, event-enhanced single-image deblurring methods, and NeRF-based image deblurring methods, which are SRN-deblur\cite{tao2018scalerecurrent}, EDI\cite{pan2019bringing}, and E$^2$NeRF\cite{qi2023e2nerf}, respectively. We compare against with these methods on the real dataset and provide the BRISQUE\cite{mittal2012no} metric. The results in \tabnref{table_quant_real} indicate a significant improvement of our method over the aforementioned methods. This is attributed to our method's incorporation of a physical model for the imaging process of blurry images, enabling better performance on real-world datasets.

\begin{table}
    \centering
    \setlength\tabcolsep{10pt}
    \caption{{\textbf{Quantitative comparisons on real-world datasets from E$^2$NeRF.}} We exploit the used real-world dataset proposed by E2NeRF \cite{qi2023e2nerf} for the evaluations, which is collected via a DAVIS event camera. The results indicates that our method outperforms EDI, SRN-Deblur and even E$^2$NeRF on the real-world datasets. Note that E$^2$NeRF \cite{qi2023e2nerf} requires multi-view images while ours only need a single image. Since E$^2$NeRF does not provide the trained model and the code for the metric computation, we re-trained E$^2$NeRF for this experiment and compute the metric with the MATLAB implementation of the BRISQUE metric for fair comparisons.}
    \resizebox{0.9\linewidth}{!}{
        \begin{tabular}{c|cccccc}
            & \multicolumn{6}{c}{BRISQUE $\downarrow$} \\
            & Camera & Lego & Letter & Plant & Toys & \textit{Average} \\
            \specialrule{0.05em}{1pt}{1pt}
            EDI\cite{pan2019bringing} & \cellcolor{orange!25}29.74 & \cellcolor{orange!25}29.35 & \cellcolor{orange!25}28.74 & \cellcolor{orange!25}31.09 & \cellcolor{orange!25}37.09 & \cellcolor{orange!25}31.20 \\
            SRN-Deblur\cite{tao2018scalerecurrent} & \cellcolor{yellow!25}32.20 & 34.91 & 40.82 & 37.45 & 46.10 & 38.30 \\
            E$^2$NeRF\cite{qi2023e2nerf} & 33.40 & \cellcolor{yellow!25}33.85 & \cellcolor{yellow!25}37.41 & \cellcolor{yellow!25}32.02 & \cellcolor{yellow!25}43.00 & \cellcolor{yellow!25}35.94\\
            
            \specialrule{0.05em}{1pt}{1pt}
            
            \textbf{BeNeRF} & \cellcolor{red!25}19.47 & \cellcolor{red!25}25.86 & \cellcolor{red!25}27.37 & \cellcolor{red!25}21.46 & \cellcolor{red!25}25.20 & \cellcolor{red!25}23.87 \\
        
        \end{tabular}
    }
	\label{table_quant_real}
\end{table}

\subsection{Qualitative evaluations}
\begin{figure*}[t]
	\setlength\tabcolsep{1pt}
	\centering
	\begin{tabular}{cccccccc}
		\raisebox{0.3in}{\rotatebox[origin=t]{90}{\scriptsize \scalebox{0.9}{Input}}} &
		\includegraphics[width=0.94\textwidth]{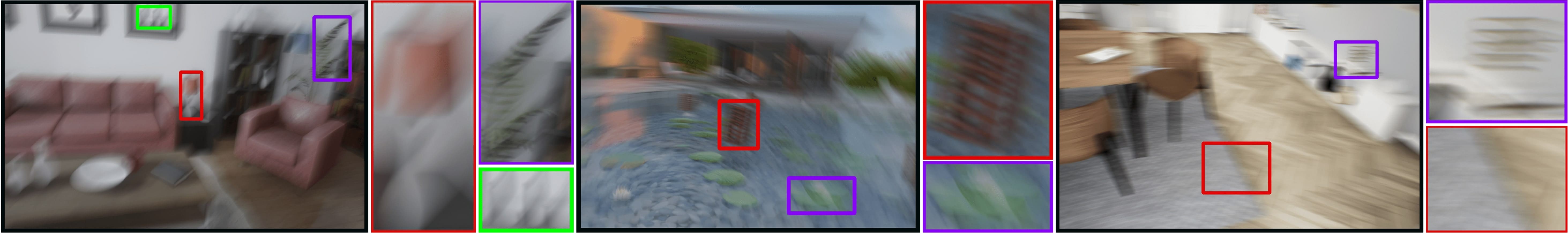}\\
		\specialrule{0em}{.02em}{.02em}
		\raisebox{0.3in}{\rotatebox[origin=t]{90}{\scriptsize \scalebox{0.9}{SRN} \cite{tao2018scalerecurrent}}} &
		\includegraphics[width=0.94\textwidth]{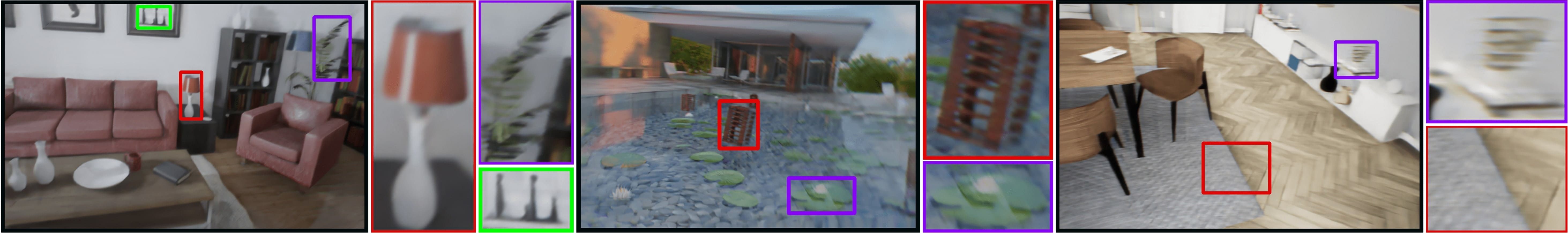}\\
		\specialrule{0em}{.02em}{.02em}
		\raisebox{0.27in}{\rotatebox[origin=t]{90}{\scriptsize \scalebox{0.9}{Restormer} \cite{zamir2022restormera}}} &
		\includegraphics[width=0.94\textwidth]{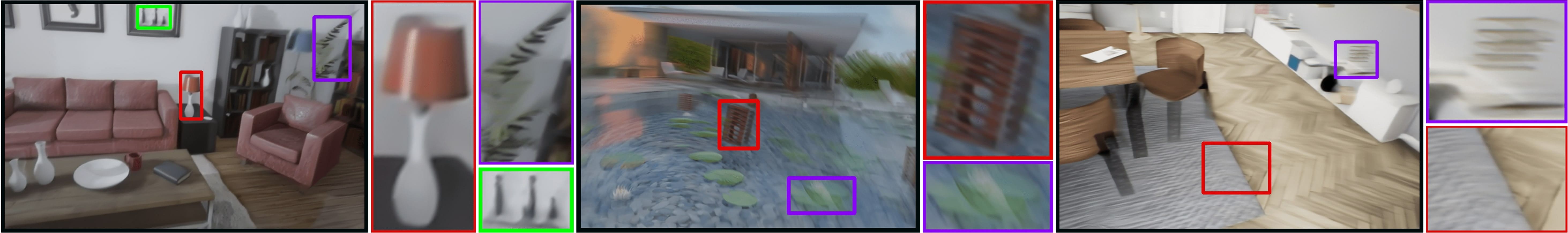}\\
		\specialrule{0em}{.02em}{.02em}
		\raisebox{0.27in}{\rotatebox[origin=t]{90}{\scriptsize \scalebox{0.9}{EDI} \cite{pan2019bringing}}} &
		\includegraphics[width=0.94\textwidth]{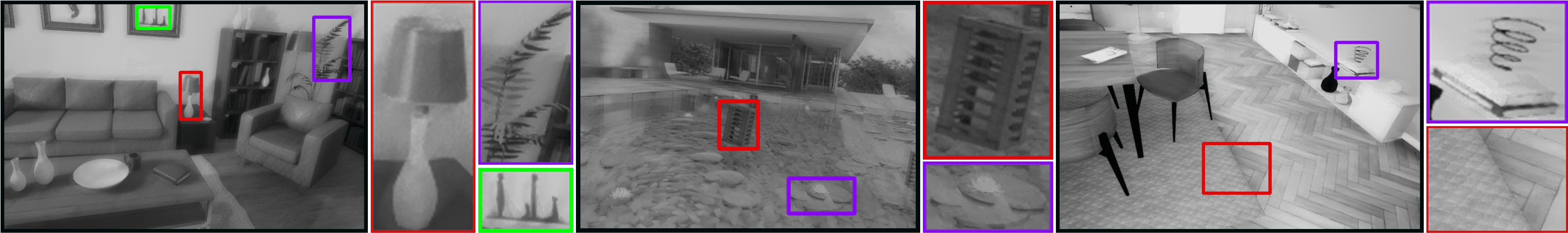}\\
		\specialrule{0em}{.02em}{.02em}
		\raisebox{0.27in}{\rotatebox[origin=t]{90}{\scriptsize \scalebox{0.9}{\textbf{BeNeRF}}}} &
		\includegraphics[width=0.94\textwidth]{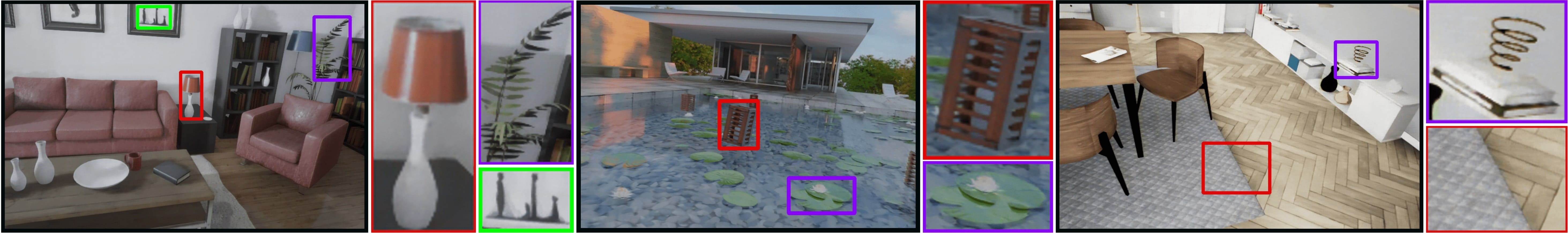}\\
		\specialrule{0em}{.02em}{.02em}
		\raisebox{0.27in}{\rotatebox[origin=t]{90}{\scriptsize \scalebox{0.9}{GT}}} &
		\includegraphics[width=0.94\textwidth]{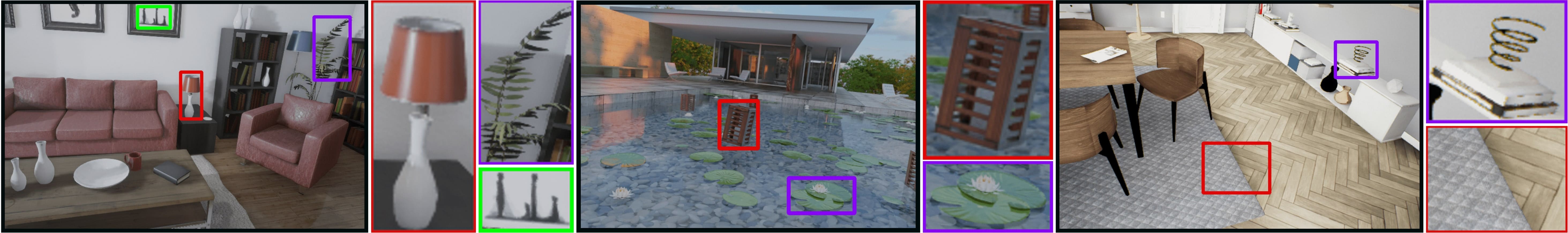}\\
		\specialrule{0em}{.02em}{.02em}
	\end{tabular}
	\caption{ {\textbf{Qualitative results of different methods with synthetic datasets.}} It demonstrates that our method delivers better performance compared to prior approaches. The learning based methods fail to generalize on severely blurry images.}
	\label{fig_synthetic}
\end{figure*}

\begin{figure*}
	\setlength\tabcolsep{2pt}
	\centering
	\begin{tabular}{ccccc}
		\includegraphics[width=0.18\textwidth]{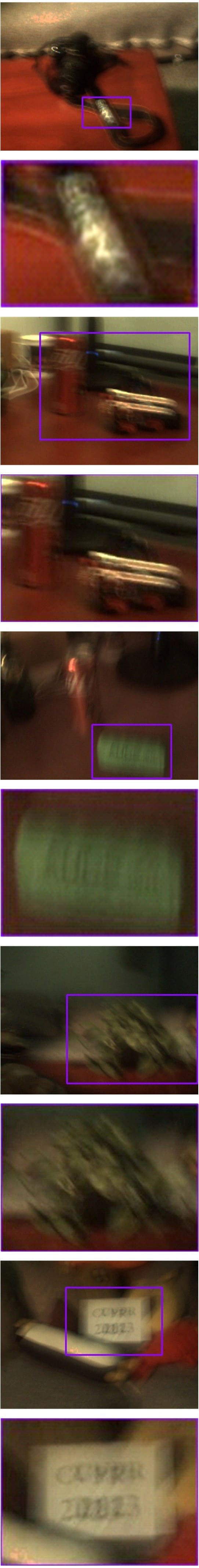} &
        \includegraphics[width=0.18\textwidth]{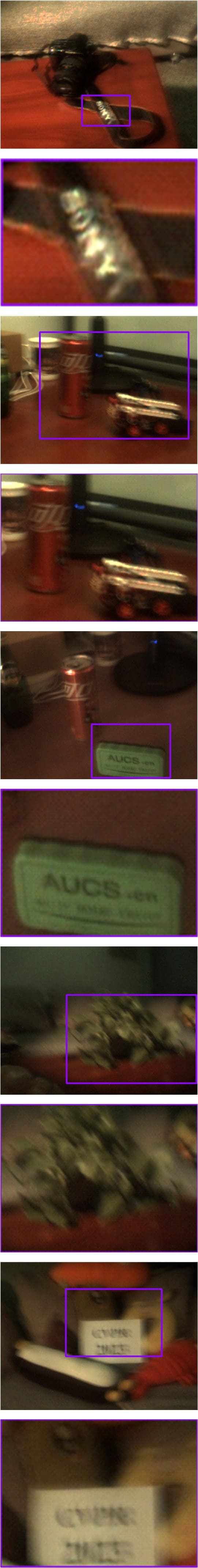} &
        \includegraphics[width=0.18\textwidth]{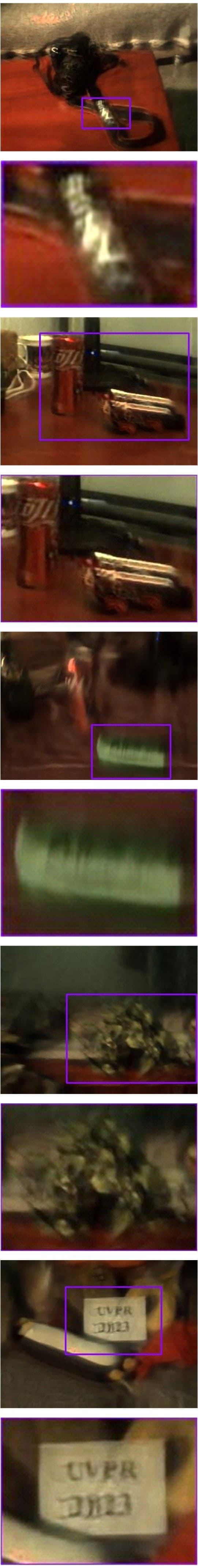} &
        \includegraphics[width=0.18\textwidth]{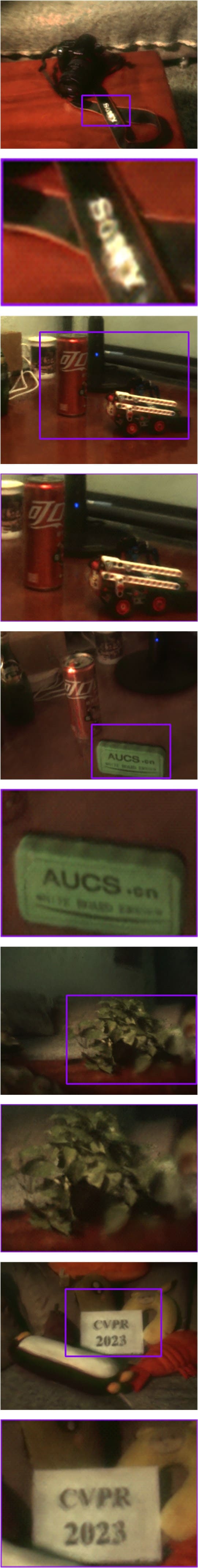} &
        \includegraphics[width=0.18\textwidth]{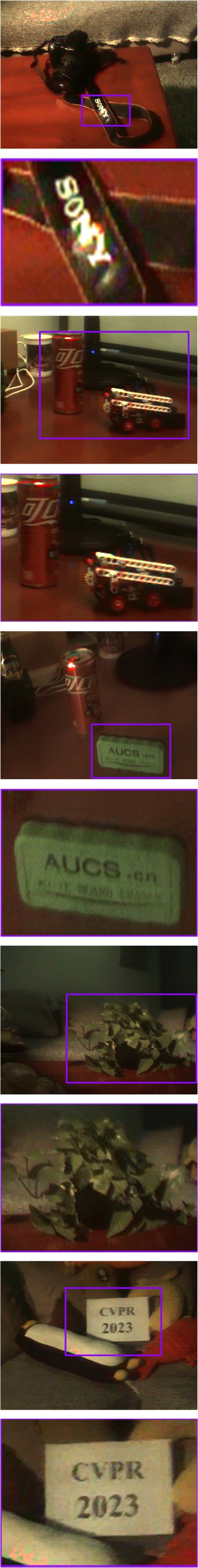} \\
		Input Image & EDI\cite{pan2019bringing} & SRN\cite{tao2018scalerecurrent} & E$^2$NeRF \cite{qi2023e2nerf} & Ours \\
	\end{tabular}
	\caption{{\textbf{Qualitative results of different methods on the real datasets.}} The experimental results demonstrate that our method delivers superior performance on the real DAVIS datasets from E$^2$NeRF. We are even achieve better performance than prior methods requiring multi-view images and longer event stream.}
	\label{fig_real}
\end{figure*}

The qualitative evaluation results are shown in \figref{fig_synthetic} and \figref{fig_real} for both synthetic and real datasets respectively. The experimental results demonstrate that our method deliver better performance than prior methods even when the image is severely blurred. In particular, \figref{fig_synthetic} shows that prior learning-based methods struggle to generalize to domain-shifted images. Notably, EDI\cite{pan2019bringing} performs well on synthetic datasets due to the high quality of event data. \figref{fig_real} shows that our method outperforms all prior methods (even trained with multi-view images) on real noisy dataset, which demonstrates the advantage of our method and the necessity to jointly optimize the camera motion and the implicit 3D representation.

\section{Conclusion}
\label{sec:con}
In conclusion, we present a novel method to jointly recover the underlying 3D scene representation and camera motion trajectory from a single blurry image and its corresponding event stream. Extensive experimental evaluations with both synthetic and real datasets demonstrate the superior performance of our method over prior works, even for those requiring multi-view images and longer event streams.

\PAR{Acknowledgements.} This work was supported in part by NSFC under Grant 62202389, in part by a grant from the Westlake University-Muyuan Joint Research Institute, and in part by the Westlake Education Foundation.

\clearpage

\title{Supplementary Material for “BeNeRF: Neural Radiance Fields from a Single Blurry Image and Event Stream”} 
\titlerunning{BeNeRF}

\author{Wenpu Li\inst{1,5}\textsuperscript{*}\orcidlink{0009-0009-6794-1810} \and
Pian Wan\inst{1,2}\textsuperscript{*}\orcidlink{0009-0007-9368-3662} \and
Peng Wang\inst{1,3}\textsuperscript{*}\orcidlink{0009-0003-5747-3319} \and
Jinghang Li\inst{4}\orcidlink{0000-0001-6196-6165} \and
Yi Zhou \inst{4}\orcidlink{0000-0003-3201-8873} \and
Peidong Liu\inst{1}\textsuperscript{$\dag$}\orcidlink{0000-0002-9767-6220}
}

\authorrunning{W. Li, P. Wan, P. Wang et al.}

\institute{\textsuperscript{1}Westlake University, \quad \textsuperscript{2}EPFL, \quad \textsuperscript{3}Zhejiang University, \quad \textsuperscript{4}Hunan University \textsuperscript{5}Guangdong University of Technology
}
\maketitle

\renewcommand{\thefootnote}{\fnsymbol{footnote}}

\appendix
\section{Introduction}

In this supplementary materials, we provide additional details and analysis on the ablation experiments. Furthermore, we present a more detailed quantitative comparison on synthetic datasets and an extensive qualitative comparison on both synthetic and real-world datasets by showcasing the deblurred images produced by various methods, facilitating a comprehensive evaluation. Finally, we provide the latent sharp videos recovered from a single blurred image using our method.

\section{More ablation studies}
\PAR{Effect of event stream.} We evaluate the effect of event stream on Livingroom dataset. Quantitative results are shown in \tabnref{table_wo_event} and qualitative results are presented in \figref{fig_wo_event}. It demonstrate that event streams can constrain the ill-posed problem caused by motion blur and only single view image as training data. Our method effectively incorporates event stream to guide NeRF to learn the correct underlying scene representation., significantly improving performance. Thus, the introduction of event streams is highly motivated.

\begin{table}
    \setlength\tabcolsep{6.5pt}
    \centering
    \caption{{\bf{Ablation studies on event stream.}} The results demonstrate that leveraging event stream can dramatically boost the performance of BeNeRF. We validate that introducing event streams is an effective method to constrain the ill-posed problem caused by motion blur and limited to a single-view image as training data.}
    \resizebox{0.6\linewidth}{!}{
        \begin{tabular}{c|ccc}
            & \multicolumn{3}{c}{Livingroom} \\
            & PSNR$\uparrow$ & SSIM$\uparrow$ & LPIPS$\downarrow$ \\
            \specialrule{0.05em}{1pt}{1pt}
            w/o event stream & 24.40 & .6612 & .4712 \\
            w/ event stream & \cellcolor{red!25}37.11 & \cellcolor{red!25}.9370 & \cellcolor{red!25}.0632 \\
         \end{tabular}
    }
    \label{table_wo_event}
\end{table}

\begin{center}
    \setlength\tabcolsep{1pt}
    \includegraphics[width=0.93\linewidth]{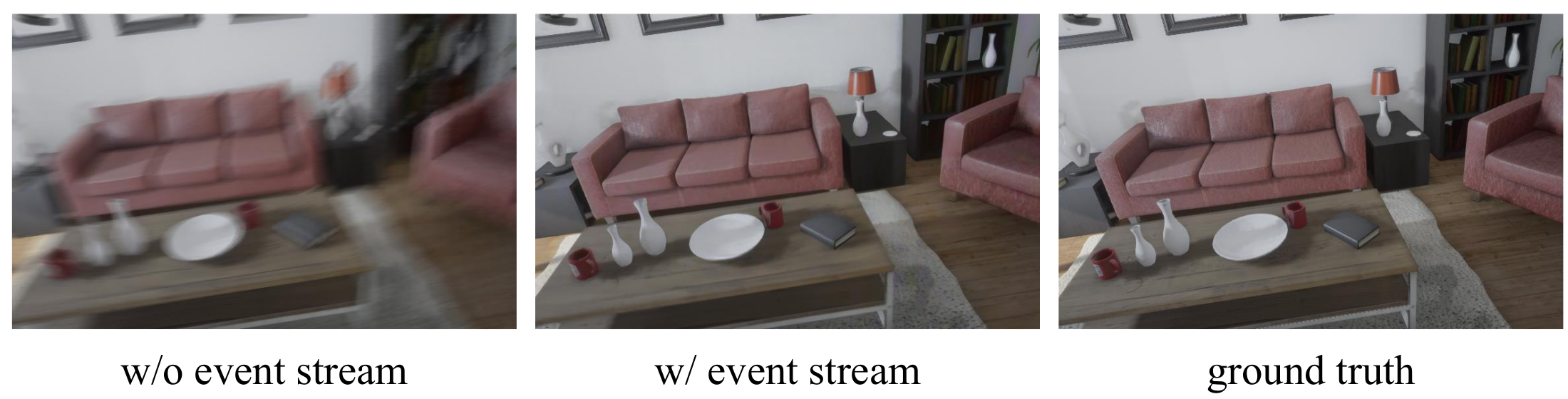}
    \captionof{figure}{\textbf{Qualitative results of ablation studies on event stream.} Here we visualize how our method benefits from event stream. The qualitative results indicate that using event stream can lead to improved model performance.}
    \label{fig_wo_event}
\end{center}

\PAR{Effect of the coarse-to-fine strategy for training.} We experiment with the effect of introducing the coarse-to-fine strategy used in BARF\cite{lin2021barf} to train our model. The results in \tabnref{table_wo_barf} show that incorporating the coarse-to-fine strategy slightly decreases model performance. It might be due that the coarse-to-fine strategy is more suit-able for multi-view case.

\begin{table}
    \setlength\tabcolsep{6.5pt}
    \centering
    \caption{{\bf{Ablation studies on coarse-to-fine strategy for training.}} The results demonstrate that using coarse-to-fine strategy to train model does not further improve the performance}
    \resizebox{0.6\linewidth}{!}{
        \begin{tabular}{c|ccc}
            & \multicolumn{3}{c}{Livingroom} \\
            & PSNR$\uparrow$ & SSIM$\uparrow$ & LPIPS$\downarrow$ \\
            \specialrule{0.05em}{1pt}{1pt}
            w/o coarse-to-fine & \cellcolor{red!25}37.11 & \cellcolor{red!25}.9370 & \cellcolor{red!25}.0632 \\
            w/ coarse-to-fine & 34.20 & .8931 & .1450 \\
        \end{tabular}
    }
    \label{table_wo_barf}
\end{table}

\section{More quantitative results}

Due to space constraints, we did not report the comparative results of SSIM metrics for each method on the synthetic dataset in the main text. In the supplementary material, we provide comprehensive quantitative comparison results, including metrics such as PSNR, SSIM, and LPIPS. The detailed quantitative results on our synthetic datasets are shown in \tabnref{table_quant_sinimage}, \tabnref{table_quant_evsinimage}, \tabnref{table_quant_nerfbased}. The experimental results presented in  \tabnref{table_quant_sinimage} demonstrate a significant superiority of our method over existing state-of-the-art single image deblurring methods. Similarly, the results in \tabnref{table_quant_evsinimage} exhibit superior performance of our method compared to event-enhanced single image deblurring methods. Additionally, the experimental results in \tabnref{table_quant_nerfbased} showcase the remarkable efficacy of our approach. Despite utilizing only a single blurred image and an event stream of a limited time interval, our method achieves performance comparable to E$^2$NeRF \cite{qi2023e2nerf}, which employs multi-view images and a longer event stream, particularly in terms of the PSNR metric. Furthermore, our method even surpasses E$^2$NeRF \cite{qi2023e2nerf} in terms of the LPIPS metric.

\begin{table}
    \centering
    \setlength\tabcolsep{1pt}
    \caption{{\textbf{Detailed quantitative comparisons on single image deblurring with synthetic datasets.}} The results demonstrate that our method significantly performs better than prior learning-based methods in terms of image quality. For HINet and NAFNet, we tests pre-trained weights from both GoPro and REDS datasets(*).}
    \resizebox{0.92\linewidth}{!}{
        \begin{tabular}{c|cccccc}
            & \multicolumn{6}{c}{PSNR $\uparrow$} \\
            & Livingroom & Whiteroom & Pinkcastle & Tanabata & Outdoorpool & \textit{Average} \\
            
            \specialrule{0.05em}{1pt}{1pt}
            
            DeblurGANv2 \cite{kupyn2019deblurganv2a} & $29.26$ & \cellcolor{yellow!25}$27.64$ & \cellcolor{orange!25}$23.16$ & \cellcolor{orange!25}$20.09$ & $26.89$ &\cellcolor{yellow!25}$25.41$ \\

            SRN-deblur \cite{tao2018scalerecurrent} &\cellcolor{orange!25}$30.86$ &$27.59$ &\cellcolor{yellow!25}$23.12$ & \cellcolor{yellow!25}$19.89$ & \cellcolor{orange!25}$27.79$ &\cellcolor{orange!25}$25.85$ \\

            MPRNet \cite{zamir2021multistage} &$28.57$ & $26.49$ & $21.60$ & $18.20$ & $27.02$ & $24.38$ \\

            HINet \cite{chen2021hinet} &$28.56$ & $26.27$ & $21.91$ & $18.59$ & $26.70$ & $24.41$ \\

            HINet* \cite{chen2021hinet} &$27.55$ & $22.89$ & $20.25$ & $18.15$ & $27.14$ & $23.20$ \\
            NAFNet \cite{chen2022simple} &\cellcolor{yellow!25}$29.92$ & \cellcolor{orange!25}$28.16$ & $22.41$ & $18.96$ & $26.75$ & $25.24$ \\
            NAFNet* \cite{chen2022simple} &$28.18$ & $23.67$ & $20.85$ & $18.38$ & \cellcolor{yellow!25}$27.52$ & $23.72$ \\
            Restormer \cite{zamir2022restormera} & $29.48$ & $27.39$ & $22.22$ & $18.82$ & $27.35$ & $25.05$ \\

            \specialrule{0.05em}{1pt}{1pt}
            
            \textbf{BeNeRF} & \cellcolor{red!25}$37.11$ & \cellcolor{red!25}$32.95$ & \cellcolor{red!25}$29.68$ & \cellcolor{red!25}$32.14$ & \cellcolor{red!25}$36.38$ & \cellcolor{red!25}$33.65$ \\
        
        \end{tabular}
    }

    \resizebox{0.92\linewidth}{!}{
        \begin{tabular}{c|cccccc}
            & \multicolumn{6}{c}{SSIM $\uparrow$} \\
            & Livingroom & Whiteroom & Pinkcastle & Tanabata & Outdoorpool & \textit{Average} \\
            
            \specialrule{0.05em}{1pt}{1pt}
            
            DeblurGANv2 \cite{kupyn2019deblurganv2a} & $.8121$ & $.7235$ & \cellcolor{orange!25}$.7043$ & \cellcolor{yellow!25}$.4964$ & $.6123$ &$.6697$ \\

            SRN-deblur \cite{tao2018scalerecurrent} &\cellcolor{orange!25}$.8437$ &\cellcolor{yellow!25}$.7396$ & \cellcolor{orange!25}$.7043$ & \cellcolor{orange!25}$.5111$ & \cellcolor{orange!25}$.6572$ & \cellcolor{orange!25}$.6912$ \\

            MPRNet \cite{zamir2021multistage} &$.7937$ & $.7301$ & $.6547$ & $.4258$ & $.6253$ & $.6459$ \\

            HINet \cite{chen2021hinet} &$.7920$ & $.6950$ & $.6625$ & $.4411$ & $.6235$ & $.6428$ \\

            HINet* \cite{chen2021hinet} &$.7822$ & $.6122$ & $.6019$ & $.4155$ & $.6211$ & $.6066$ \\
            NAFNet \cite{chen2022simple} & \cellcolor{yellow!25}$.8306$ & \cellcolor{orange!25}$.7874$ & \cellcolor{yellow!25}$.6896$ & $.4665$ & $.6255$ & \cellcolor{yellow!25}$.6799$ \\
            NAFNet* \cite{chen2022simple} &$.7991$ & $.6422$ & $.6175$ & $.4230$ & \cellcolor{yellow!25}$.6407$ & $.6245$ \\
            Restormer \cite{zamir2022restormera} & $.8262$ & $.7314$ & $.6803$ & $.4596$ & $.6352$ & $.6665$ \\

            \specialrule{0.05em}{1pt}{1pt}
            
            \textbf{BeNeRF} & \cellcolor{red!25}$.9370$ & \cellcolor{red!25}$.8651$ & \cellcolor{red!25}$.8593$ & \cellcolor{red!25}$.9015$ & \cellcolor{red!25}$.9039$ & \cellcolor{red!25}$.8934$ \\
        
        \end{tabular}
    }

    \resizebox{0.92\linewidth}{!}{
        \begin{tabular}{c|cccccc}
            & \multicolumn{6}{c}{LPIPS $\downarrow$} \\
            & Livingroom & Whiteroom & Pinkcastle & Tanabata & Outdoorpool & \textit{Average} \\
            \specialrule{0.05em}{1pt}{1pt}
            
            DeblurGANv2 \cite{kupyn2019deblurganv2a} &\cellcolor{orange!25}.2087 &\cellcolor{orange!25}.1989 &\cellcolor{orange!25}.2608 &\cellcolor{yellow!25}.3934 &\cellcolor{orange!25}.3100 &\cellcolor{orange!25}.2744 \\

            SRN-deblur \cite{tao2018scalerecurrent} &.2529 &.2503 &.3245 & .4260 & .3594 & .3226 \\

            MPRNet \cite{zamir2021multistage} & .2621 & .2564 & .3586 & .4173 & .3679 & .3325 \\

            HINet \cite{chen2021hinet} & .2468 & .2620 & .3500 & .4024 & .3355 & .3193 \\

            HINet* \cite{chen2021hinet} & .3327 & .3602 & .3789 & .5265 & .4397 & .4076 \\
            NAFNet \cite{chen2022simple} & \cellcolor{yellow!25}.2268 &\cellcolor{yellow!25} .1991 & \cellcolor{yellow!25}.3058 & \cellcolor{orange!25}.3908 & \cellcolor{yellow!25}.3280 & \cellcolor{yellow!25}.2901 \\
            NAFNet* \cite{chen2022simple} & .3182 & .3566 & .3943 & .5271 & .4257 & .4044 \\
            Restormer \cite{zamir2022restormera} & .2391 & .2493 & .3373 & .4248 & .3664 & .3234 \\

            \specialrule{0.05em}{1pt}{1pt}
            
            \textbf{BeNeRF} &\cellcolor{red!25}.0632 &\cellcolor{red!25}.0788 &\cellcolor{red!25}.0761 &\cellcolor{red!25}.0515 &\cellcolor{red!25}.0677 &\cellcolor{red!25}.0675 \\
        
        \end{tabular}
    }
	\label{table_quant_sinimage}
\end{table}

\begin{table}
    \centering
    \caption{{\textbf{Detailed quantitative comparisons on event-enhanced single image deblurring with synthetic datasets.}} The results demonstrate that our method performs better than both EDI and eSLNet.}
    \resizebox{0.83\linewidth}{!}{
        \begin{tabular}{c|cccccc}
            & \multicolumn{6}{c}{PSNR $\uparrow$} \\
            & Livingroom & Whiteroom & Pinkcastle & Tanabata & Outdoorpool & \textit{Average} \\
            \specialrule{0.05em}{1pt}{1pt}
            
            eSLNet\cite{wang2020event} & \cellcolor{yellow!25}14.22 & \cellcolor{yellow!25}10.81 & \cellcolor{yellow!25}10.49 & \cellcolor{yellow!25}8.86 & \cellcolor{yellow!25}11.80 & \cellcolor{yellow!25}11.24 \\
            EDI\cite{pan2019bringing} & \cellcolor{orange!25}32.61 & \cellcolor{orange!25}30.33 & \cellcolor{orange!25}27.24 & \cellcolor{orange!25}24.87 & \cellcolor{orange!25}31.64 & \cellcolor{orange!25}29.34 \\
            
            \specialrule{0.05em}{1pt}{1pt}
            
            \textbf{BeNeRF} &\cellcolor{red!25}37.11 &\cellcolor{red!25}32.95 &\cellcolor{red!25}29.68 &\cellcolor{red!25}32.14 &\cellcolor{red!25}36.38 &\cellcolor{red!25}33.65 \\
        
        \end{tabular}
    }

    \resizebox{0.83\linewidth}{!}{
        \begin{tabular}{c|cccccc}
            & \multicolumn{6}{c}{SSIM $\uparrow$} \\
            & Livingroom & Whiteroom & Pinkcastle & Tanabata & Outdoorpool & \textit{Average} \\
            \specialrule{0.05em}{1pt}{1pt}
            
            eSLNet\cite{wang2020event} & \cellcolor{yellow!25}.3527 & \cellcolor{yellow!25}.2156 & \cellcolor{yellow!25}.2903 & \cellcolor{yellow!25}.1658 & \cellcolor{yellow!25}.2181 & \cellcolor{yellow!25}.2485 \\
            EDI\cite{pan2019bringing} & \cellcolor{orange!25}.8871 & \cellcolor{orange!25}.8152 & \cellcolor{orange!25}.8356 & \cellcolor{orange!25}.7564 & \cellcolor{orange!25}.8044 & \cellcolor{orange!25}.8197 \\
            
            \specialrule{0.05em}{1pt}{1pt}
            
            \textbf{BeNeRF} &\cellcolor{red!25}.9370 &\cellcolor{red!25}.8651 &\cellcolor{red!25}.8593 &\cellcolor{red!25}.9015 &\cellcolor{red!25}.9039 &\cellcolor{red!25}.8934 \\
        
        \end{tabular}
    }
    
    \resizebox{0.83\linewidth}{!}{
        \begin{tabular}{c|cccccc}
            & \multicolumn{6}{c}{LPIPS $\downarrow$} \\
            & Livingroom & Whiteroom & Pinkcastle & Tanabata & Outdoorpool & \textit{Average} \\
            \specialrule{0.05em}{1pt}{1pt}
            
            eSLNet\cite{wang2020event} & \cellcolor{yellow!25}.3981 & \cellcolor{yellow!25}.4236 & \cellcolor{yellow!25}.4902 & \cellcolor{yellow!25}.5067 & \cellcolor{yellow!25}.4676 & \cellcolor{yellow!25}.4572 \\
            EDI\cite{pan2019bringing} & \cellcolor{orange!25}.0904 & \cellcolor{orange!25}.1020 & \cellcolor{orange!25}.0779 & \cellcolor{orange!25}.1039 & \cellcolor{orange!25}.1409 & \cellcolor{orange!25}.1030 \\

            \specialrule{0.05em}{1pt}{1pt}
            
            \textbf{BeNeRF} &\cellcolor{red!25}.0632 &\cellcolor{red!25}.0788 &\cellcolor{red!25}.0761 &\cellcolor{red!25}.0515 &\cellcolor{red!25}.0677 &\cellcolor{red!25}.0675 \\
        
        \end{tabular}
    }
	\label{table_quant_evsinimage}
\end{table}

\begin{table}
    \centering
    \caption{{\textbf{Detailed quantitative comparisons on NeRF-based image deblurring with synthetic datasets from E$^2$NeRF.}} The results indicate that our method outperforms both NeRF and Deblur-NeRF, and exhibits performance comparable to E$^2$NeRF in terms of the PSNR metric. Moreover, our method even surpasses E$^2$NeRF with the LPIPS metric.}
    \setlength\tabcolsep{4pt}
    \resizebox{0.83\linewidth}{!}{
        \begin{tabular}{c|ccccccc}
            & \multicolumn{7}{c}{PSNR $\uparrow$} \\
            & Chair & Ficus & Hotdog & Lego & Materials & Mic & \textit{Average} \\
            \specialrule{0.05em}{1pt}{1pt}

            NeRF\cite{mildenhall2020nerf} & 24.29 & \cellcolor{yellow!25}22.98 & \cellcolor{yellow!25}27.75 & 21.95 & 19.99 & \cellcolor{yellow!25}20.50 & \cellcolor{yellow!25}22.91 \\
            Deblur-NeRF\cite{ma2022deblurnerf} & \cellcolor{yellow!25}25.87 & 22.86 & 24.62 & \cellcolor{yellow!25}24.47 & \cellcolor{yellow!25}20.54 & 11.92 & 21.71 \\
            E$^2$NeRF\cite{qi2023e2nerf} & \cellcolor{red!25}31.28 &\cellcolor{orange!25}30.00 & \cellcolor{red!25}34.34 & \cellcolor{red!25}28.11 & \cellcolor{orange!25}27.27 & \cellcolor{red!25}27.60 & \cellcolor{red!25}29.77 \\
            
            \specialrule{0.05em}{1pt}{1pt}
            
            \textbf{BeNeRF} &\cellcolor{orange!25}31.17 &\cellcolor{red!25}30.81 &\cellcolor{orange!25}34.31 &\cellcolor{orange!25}28.09 &\cellcolor{red!25}27.44 &\cellcolor{orange!25}26.13 & \cellcolor{orange!25}29.66\\
        
        \end{tabular}
    }

    \resizebox{0.83\linewidth}{!}{
        \begin{tabular}{c|ccccccc}
            & \multicolumn{7}{c}{SSIM $\uparrow$} \\
            & Chair & Ficus & Hotdog & Lego & Materials & Mic & \textit{Average} \\
            \specialrule{0.05em}{1pt}{1pt}

            NeRF\cite{mildenhall2020nerf} & .9357 & \cellcolor{yellow!25}.9023 & \cellcolor{orange!25}.9546 & .8548 & \cellcolor{yellow!25}.9108 & \cellcolor{yellow!25}.8854 & \cellcolor{yellow!25}.9072 \\
            
            Deblur-NeRF\cite{ma2022deblurnerf} & \cellcolor{yellow!25}.9373 & .8982 & .9396 & \cellcolor{yellow!25}.8756 & .9012 & .7249 & .8795 \\
            
            E$^2$NeRF\cite{qi2023e2nerf} & \cellcolor{red!25}.9749 & \cellcolor{red!25}.9663 & \cellcolor{red!25}.9784 & \cellcolor{red!25}.9339 & \cellcolor{red!25}.9570 & \cellcolor{red!25}.9496 & \cellcolor{red!25}.9600 \\
            
            \specialrule{0.05em}{1pt}{1pt}
            
            \textbf{BeNeRF} &\cellcolor{orange!25}.9488 &\cellcolor{orange!25} .9465 &\cellcolor{yellow!25}.9497 &\cellcolor{orange!25}.8930 &\cellcolor{orange!25}.9144 &\cellcolor{orange!25}.9115 & \cellcolor{orange!25}.9273 \\
        
        \end{tabular}
    }
    
    \resizebox{0.83\linewidth}{!}{
        \begin{tabular}{c|ccccccc}
            & \multicolumn{7}{c}{LPIPS $\downarrow$} \\
            & Chair & Ficus & Hotdog & Lego & Materials & Mic & \textit{Average} \\
            \specialrule{0.05em}{1pt}{1pt}

            NeRF\cite{mildenhall2020nerf} & \cellcolor{yellow!25}.1254 & \cellcolor{yellow!25}.1037 & \cellcolor{yellow!25}.1158 & .2103 & \cellcolor{yellow!25}.1512 & \cellcolor{yellow!25}.1579 & \cellcolor{yellow!25}.1441 \\
            Deblur-NeRF\cite{ma2022deblurnerf} & .2185 & .1541 & .2138 & \cellcolor{yellow!25}.2053 & .2562 & .3706 & .2364 \\
            E$^2$NeRF\cite{qi2023e2nerf} & \cellcolor{orange!25}.0608 & \cellcolor{orange!25}.0362 & \cellcolor{orange!25}.0660 & \cellcolor{orange!25}.1078 & \cellcolor{orange!25}.0919 & \cellcolor{red!25}.0724 & \cellcolor{orange!25}.0725 \\
            
            \specialrule{0.05em}{1pt}{1pt}
            
            \textbf{BeNeRF} &\cellcolor{red!25}.0500 &\cellcolor{red!25} .0299 &\cellcolor{red!25}.0539 &\cellcolor{red!25}.0745 &\cellcolor{red!25}.0708 &\cellcolor{orange!25}.0738 & \cellcolor{red!25}.0588 \\
        
        \end{tabular}
    }
	\label{table_quant_nerfbased}
\end{table}

\section{More qualitative results}

\begin{figure*}[t]
	\centering
	\begin{tabular}{cccccccc}
		\raisebox{0.25in}{\rotatebox[origin=t]{90}{\scriptsize \scalebox{0.9}{Input}}} &
		\includegraphics[width=1\textwidth,height=0.6in]{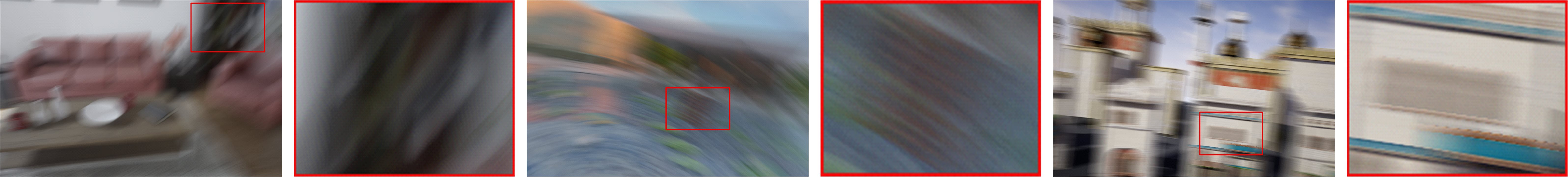}\\
		\specialrule{0em}{.02em}{.02em}
		\raisebox{0.25in}{\rotatebox[origin=t]{90}{\scriptsize \scalebox{0.8}{DeblurGANv2} }} &
		\includegraphics[width=1\textwidth,height=0.6in]{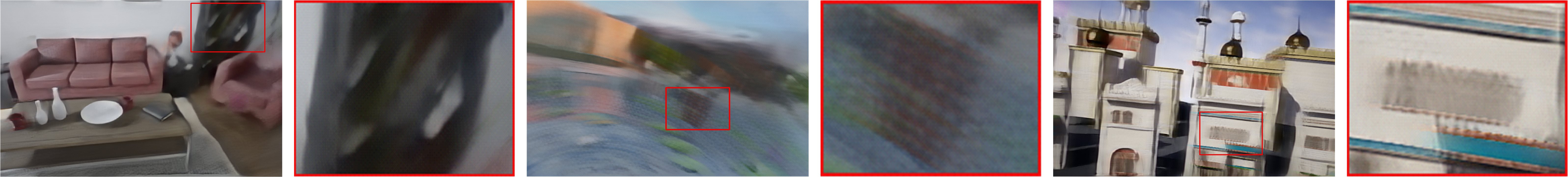}\\
		\specialrule{0em}{.02em}{.02em}
		\raisebox{0.27in}{\rotatebox[origin=t]{90}{\scriptsize \scalebox{0.9}{SRN}}} &
		\includegraphics[width=1\textwidth,height=0.6in]{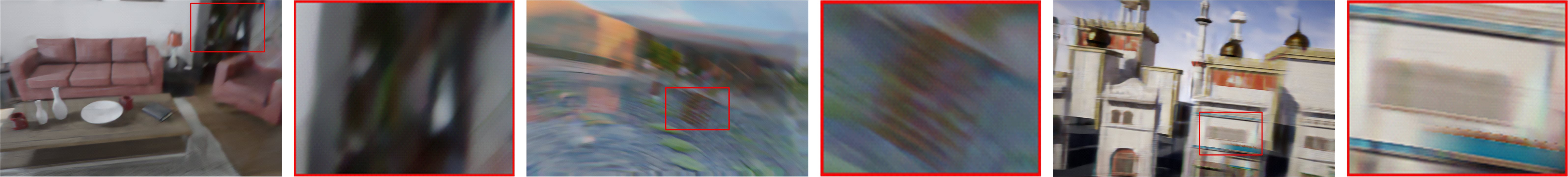}\\
		\specialrule{0em}{.02em}{.02em}
		\raisebox{0.27in}{\rotatebox[origin=t]{90}{\scriptsize \scalebox{0.9}{NAFNet} }} &
		\includegraphics[width=1\textwidth,height=0.6in]{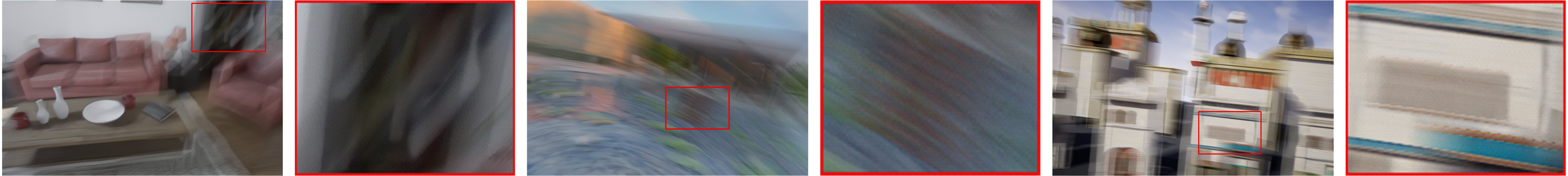}\\
		\specialrule{0em}{.02em}{.02em}
		\raisebox{0.27in}{\rotatebox[origin=t]{90}{\scriptsize \scalebox{0.9}{Restormer}}} &
		\includegraphics[width=1\textwidth,height=0.6in]{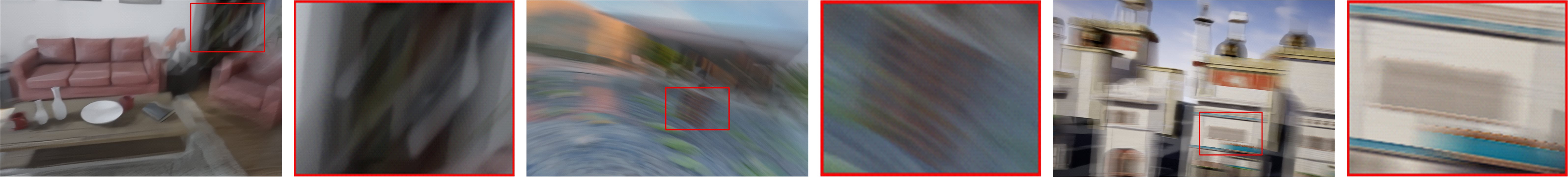}\\
		\specialrule{0em}{.02em}{.02em}
		\raisebox{0.27in}{\rotatebox[origin=t]{90}{\scriptsize \scalebox{0.9}{EDI}}} &
		\includegraphics[width=1\textwidth,height=0.6in]{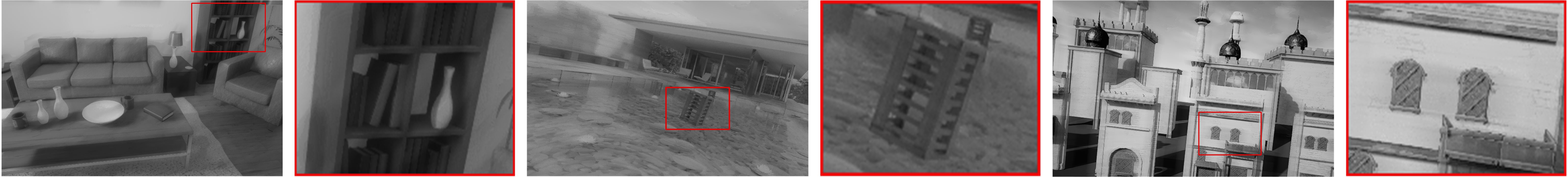}\\
		\specialrule{0em}{.02em}{.02em}
        \raisebox{0.27in}{\rotatebox[origin=t]{90}{\scriptsize \scalebox{0.9}{\textbf{BeNeRF}}}} &
		\includegraphics[width=1\textwidth,height=0.6in]{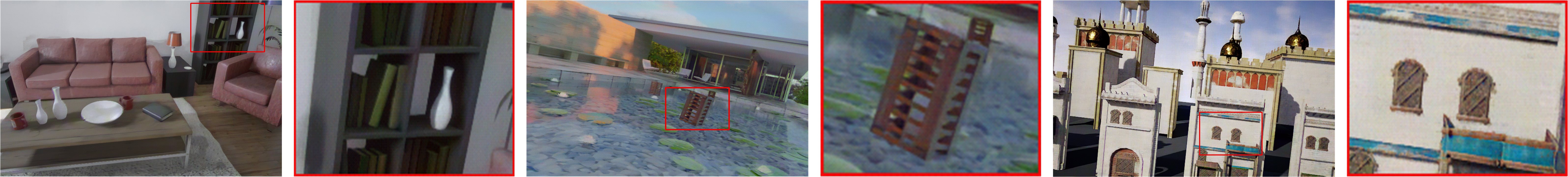}\\
		\specialrule{0em}{.02em}{.02em}
  		\raisebox{0.27in}{\rotatebox[origin=t]{90}{\scriptsize \scalebox{0.9}{GT}}} &
		\includegraphics[width=1\textwidth,height=0.6in]{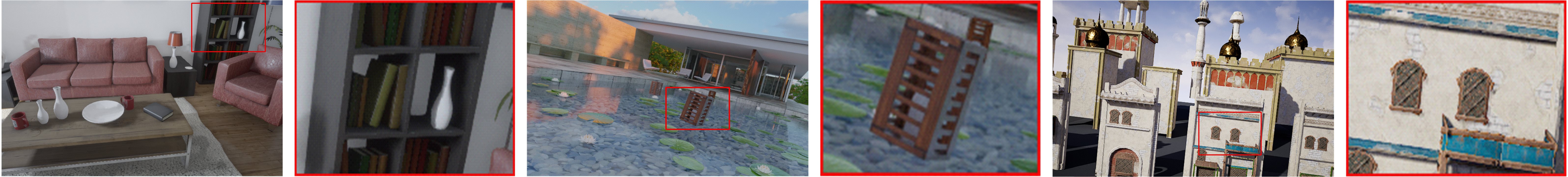}\\
		\specialrule{0em}{.02em}{.02em}
	\end{tabular}
	\caption{ {\textbf{Qualitative results of different methods with synthetic datasets.}} Detailed qualitative comparison for “Livingroom”, "Outdoorpool" and "Pinkcastle" scene of synthetic dataset.}
	\label{fig_our_synthetic_1}
\end{figure*}

\begin{figure*}[t]
	\centering
	\begin{tabular}{cccccccc}
		\raisebox{0.35in}{\rotatebox[origin=t]{90}{\scriptsize \scalebox{0.9}{Input}}} &
		\includegraphics[width=1\textwidth,height=0.82in]{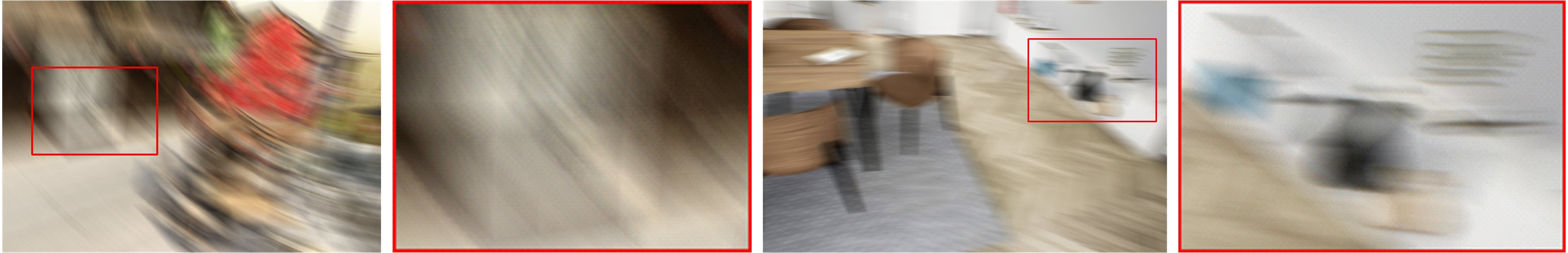}\\
		\specialrule{0em}{.02em}{.02em}
		\raisebox{0.35in}{\rotatebox[origin=t]{90}{\scriptsize \scalebox{0.8}{DeblurGANv2} }} &
		\includegraphics[width=1\textwidth,height=0.82in]{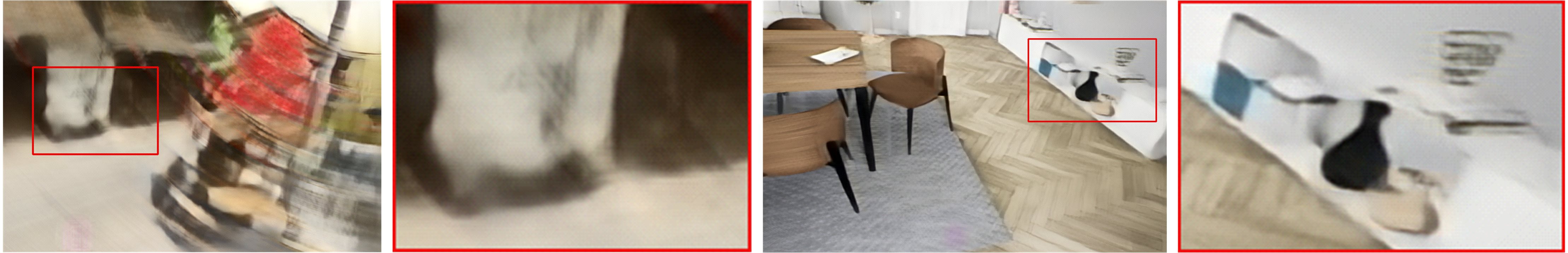}\\
		\specialrule{0em}{.02em}{.02em}
		\raisebox{0.35in}{\rotatebox[origin=t]{90}{\scriptsize \scalebox{0.9}{SRN}}} &
		\includegraphics[width=1\textwidth,height=0.82in]{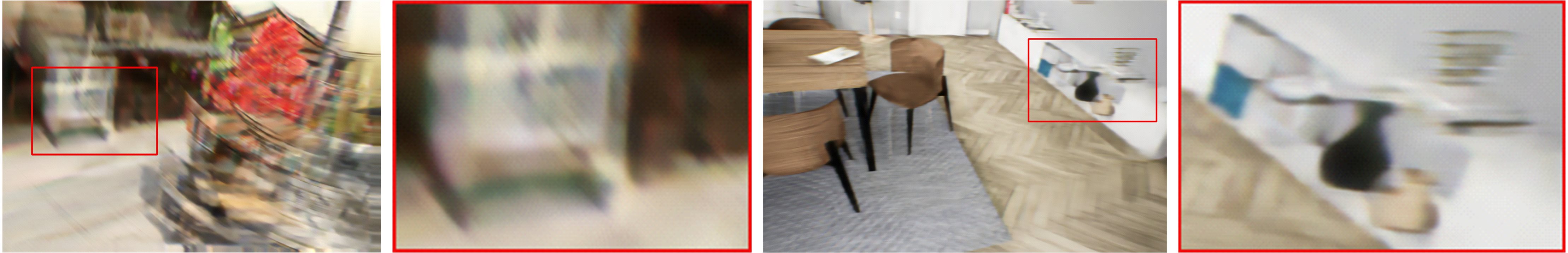}\\
		\specialrule{0em}{.02em}{.02em}
		\raisebox{0.35in}{\rotatebox[origin=t]{90}{\scriptsize \scalebox{0.9}{NAFNet} }} &
		\includegraphics[width=1\textwidth,height=0.82in]{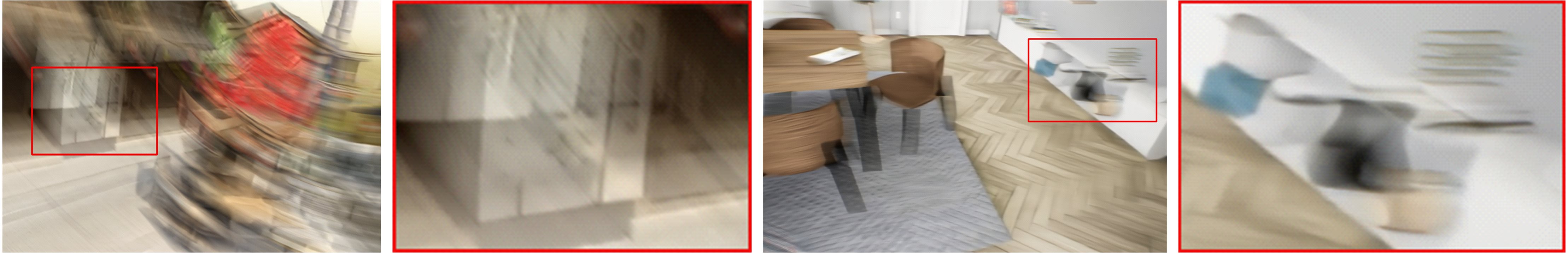}\\
		\specialrule{0em}{.02em}{.02em}
		\raisebox{0.35in}{\rotatebox[origin=t]{90}{\scriptsize \scalebox{0.9}{Restormer}}} &
		\includegraphics[width=1\textwidth,height=0.82in]{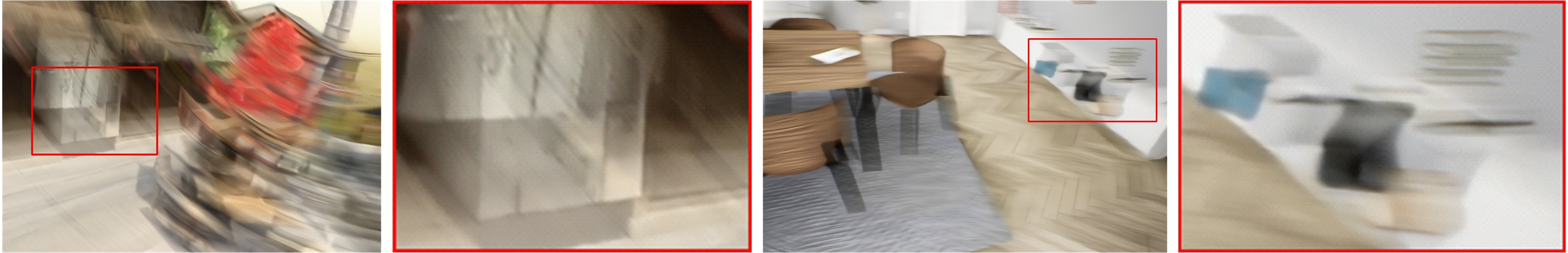}\\
		\specialrule{0em}{.02em}{.02em}
		\raisebox{0.35in}{\rotatebox[origin=t]{90}{\scriptsize \scalebox{0.9}{EDI}}} &
		\includegraphics[width=1\textwidth,height=0.82in]{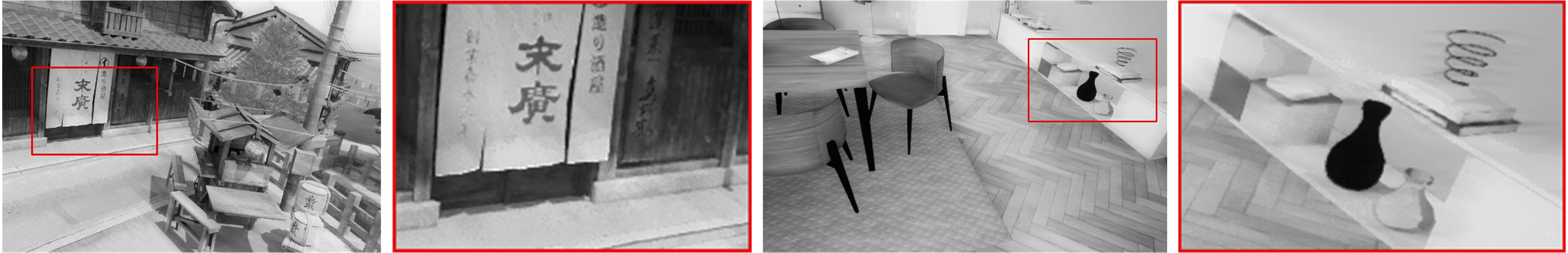}\\
		\specialrule{0em}{.02em}{.02em}
        \raisebox{0.35in}{\rotatebox[origin=t]{90}{\scriptsize \scalebox{0.9}{\textbf{BeNeRF}}}} &
		\includegraphics[width=1\textwidth,height=0.82in]{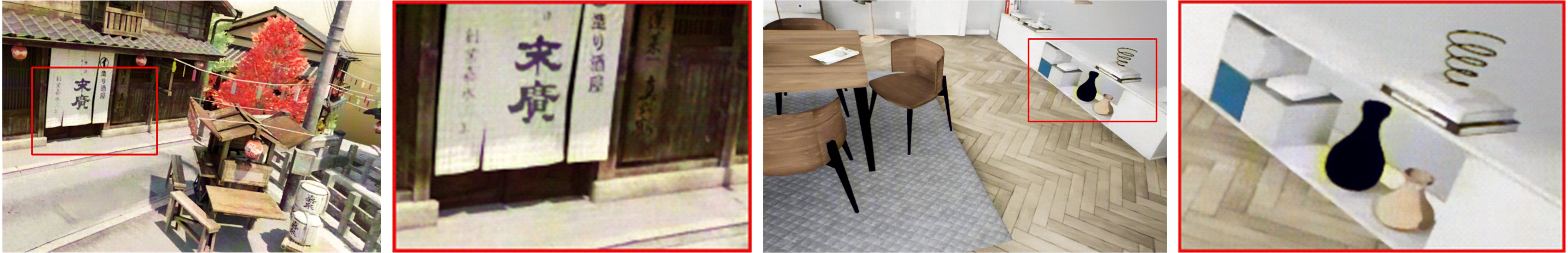}\\
		\specialrule{0em}{.02em}{.02em}
  		\raisebox{0.35in}{\rotatebox[origin=t]{90}{\scriptsize \scalebox{0.9}{GT}}} &
		\includegraphics[width=1\textwidth,height=0.82in]{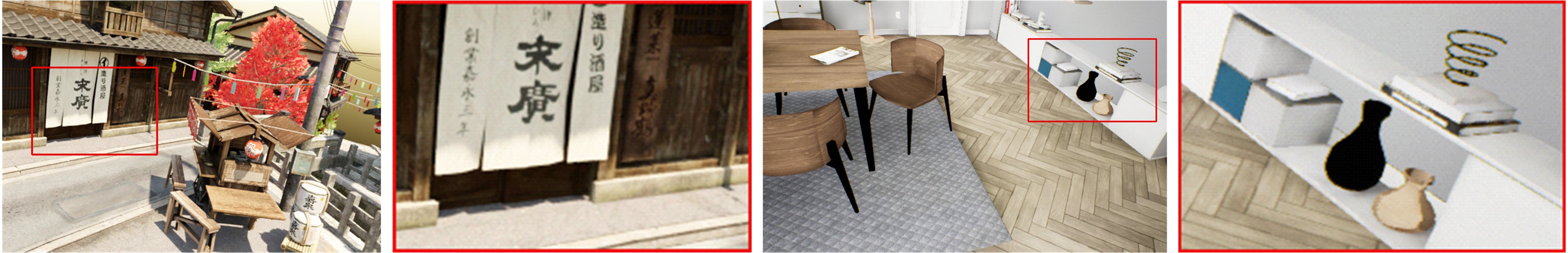}\\
		\specialrule{0em}{.02em}{.02em}
	\end{tabular}
	\caption{ {\textbf{Qualitative results of different methods with synthetic datasets.}} Detailed qualitative comparison for “Tanabata” and "Whiteroom" scene of synthetic dataset.}
	\label{fig_our_synthetic_2}
\end{figure*}

We conducted a detailed comparison of our proposed method with single-image deblurring methods and event-enhanced single-image deblurring methods on our proposed synthetic dataset, consisting of six scenes(\ie Livingroom, Whiteroom, Pinkcastle, Tanabata and Outdoorpool). The motion blur images are synthesized by importing real camera motion trajectories from ETH3D \cite{schops2019bad}. Thorough qualitative comparison in \figref{fig_our_synthetic_1} and \figref{fig_our_synthetic_2} illustrate that even under severe motion blur conditions, our method can effectively recover sharp images, demonstrating significant superiority over existing state-of-the-art single-image deblurring methods. Since the event stream synthesized using ESIM \cite{rebecq2018esim} is monochannel, the images recovered using EDI\cite{pan2019bringing} are grayscale.

\begin{figure*}
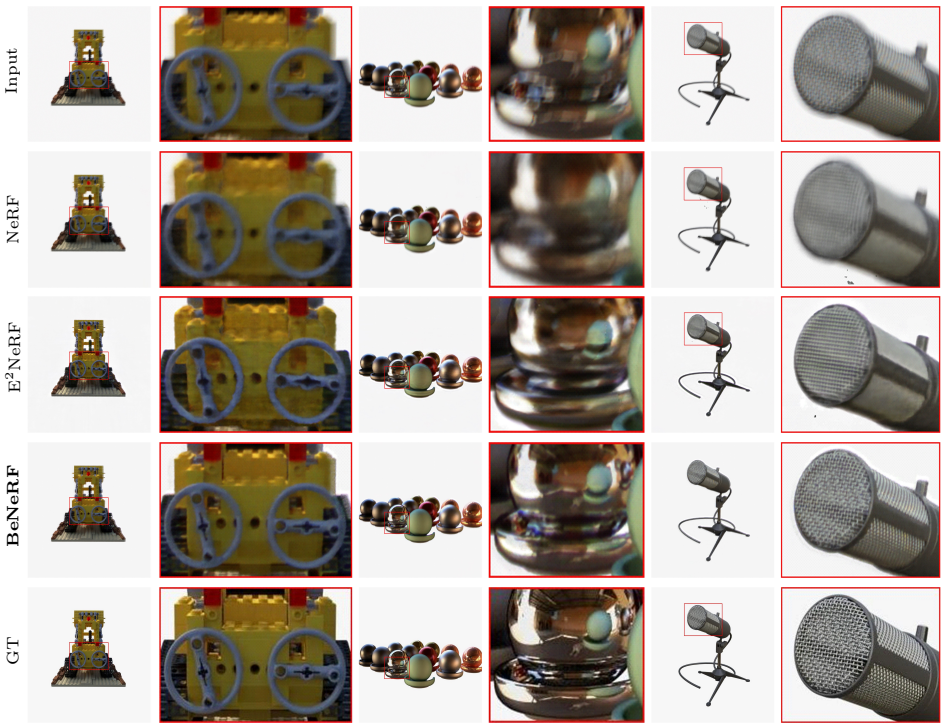

	\centering
	\begin{tabular}{cccccccc}
		\raisebox{0.31in}{\rotatebox[origin=t]{90}{\scriptsize \scalebox{0.9}{Input}}} &
		\includegraphics[width=1\textwidth,height=0.7in]{fig/supplementary/e2nerf_synthetic_datasets/46-1.jpg}\\
		\specialrule{0em}{.02em}{.02em}
		\raisebox{0.31in}{\rotatebox[origin=t]{90}{\scriptsize \scalebox{0.9}{NeRF} }} &
		\includegraphics[width=1\textwidth,height=0.7in]{fig/supplementary/e2nerf_synthetic_datasets/46-2.jpg}\\
		\specialrule{0em}{.02em}{.02em}
		\raisebox{0.31in}{\rotatebox[origin=t]{90}{\scriptsize \scalebox{0.9}{E$^2$NeRF}}} &
		\includegraphics[width=1\textwidth,height=0.7in]{fig/supplementary/e2nerf_synthetic_datasets/46-4.jpg}\\
		\specialrule{0em}{.02em}{.02em}
		\raisebox{0.31in}{\rotatebox[origin=t]{90}{\scriptsize \scalebox{0.9}{\textbf{BeNeRF}}}} &
		\includegraphics[width=1\textwidth,height=0.7in]{fig/supplementary/e2nerf_synthetic_datasets/46-5.jpg}\\
		\specialrule{0em}{.02em}{.02em}
		\raisebox{0.31in}{\rotatebox[origin=t]{90}{\scriptsize \scalebox{0.9}{GT}}} &
		\includegraphics[width=1\textwidth,height=0.7in]{fig/supplementary/e2nerf_synthetic_datasets/46-6.jpg}\\
		\specialrule{0em}{.02em}{.02em}
	\end{tabular}
	\caption{ {\textbf{Qualitative results of different methods with synthetic datasets proposed by E$^2$NeRF.}} Detailed qualitative comparison for “Lego”, "Materials" and "Mic" scene of synthetic dataset from E$^2$NeRF.}
	\label{fig_e2nerf_synthetic_2}
\end{figure*}

To facilitate a comparative analysis with a NeRF-based image deblurring method leveraging multi-view information, we conducted a detailed evaluation on the synthetic dataset proposed by E$^2$NeRF\cite{qi2023e2nerf}, which comprises six distinct scenes(\ie Chair, Ficus, Hotdog, Lego, Materials, and Mic). Extensive qualitative comparisons in \figref{fig_e2nerf_synthetic_1} and \figref{fig_e2nerf_synthetic_2} demonstrate the superiority of our method, despite utilizing only a single blurred image and a short event stream, over methods that utilize multi-view images and long event streams.

\begin{figure*}[t]
	\centering
	\begin{tabular}{cccccccc}
		\raisebox{0.4in}{\rotatebox[origin=t]{90}{\scriptsize \scalebox{1.2}{Input}}} &
		\includegraphics[width=1\textwidth,height=1in]{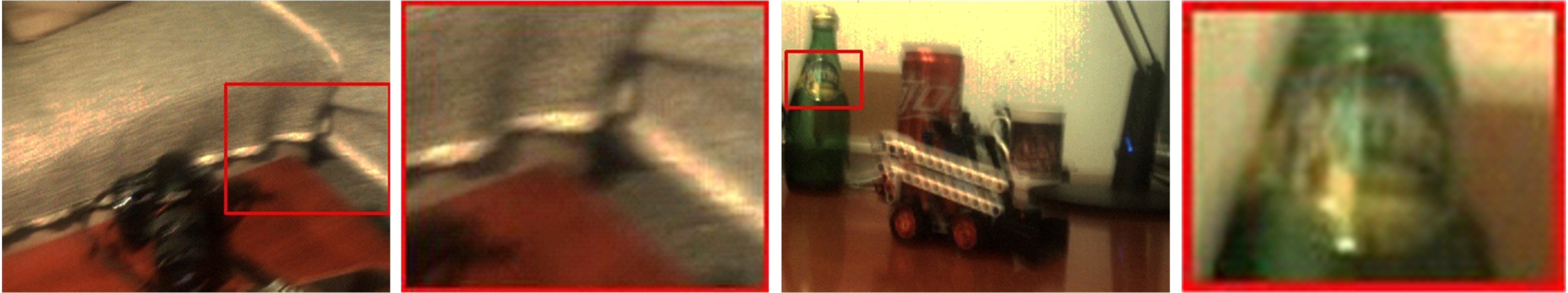}\\
		\specialrule{0em}{.02em}{.02em}
		\raisebox{0.4in}{\rotatebox[origin=t]{90}{\scriptsize \scalebox{1.2}{SRN} }} &
		\includegraphics[width=1\textwidth,height=1in]{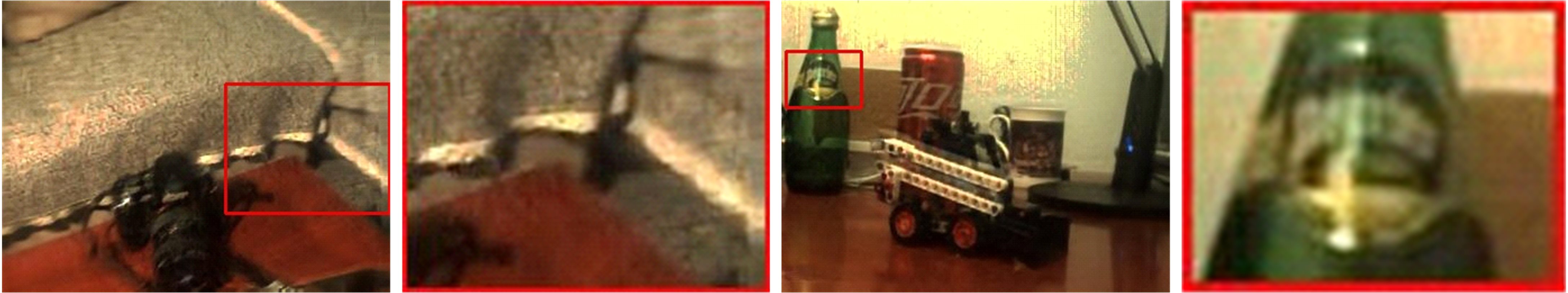}\\
		\specialrule{0em}{.02em}{.02em}
		\raisebox{0.4in}{\rotatebox[origin=t]{90}{\scriptsize \scalebox{1.2}{EDI}}} &
		\includegraphics[width=1\textwidth,height=1in]{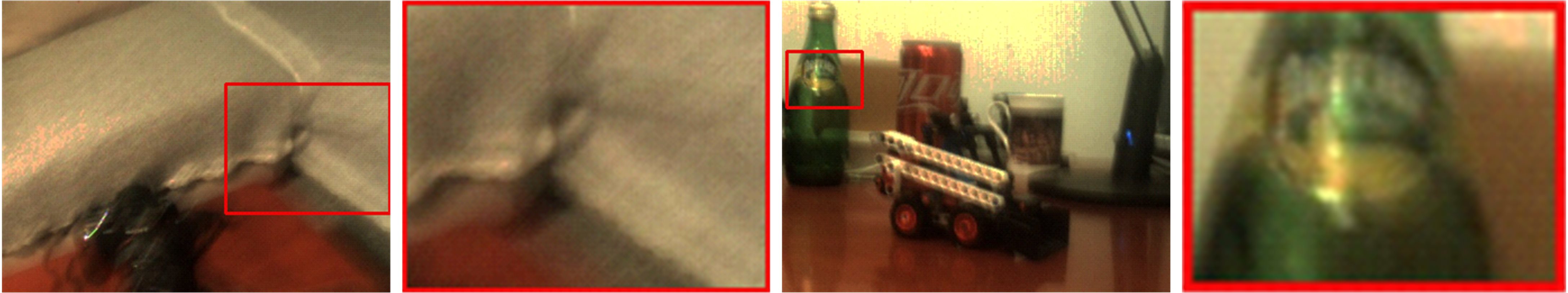}\\
		\specialrule{0em}{.02em}{.02em}
		\raisebox{0.4in}{\rotatebox[origin=t]{90}{\scriptsize \scalebox{1.2}{E$^2$NeRF}}} &
		\includegraphics[width=1\textwidth,height=1in]{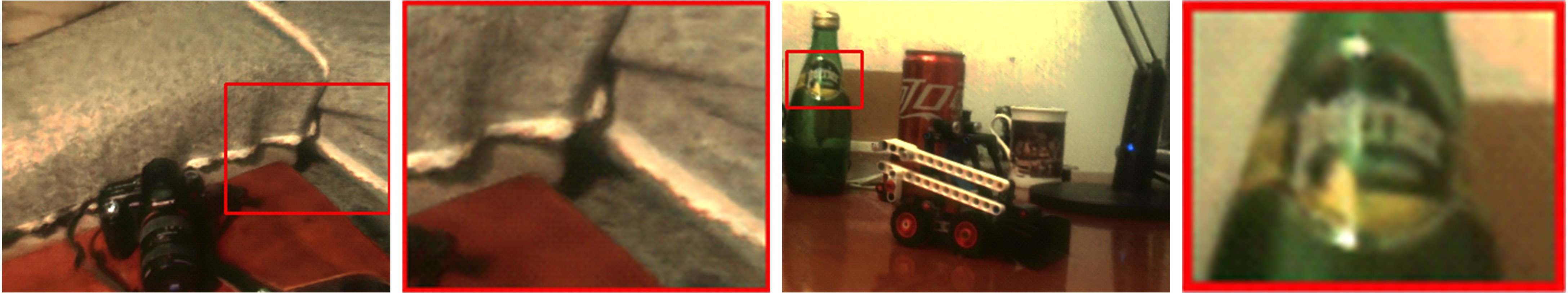}\\
		\specialrule{0em}{.02em}{.02em}
		\raisebox{0.4in}{\rotatebox[origin=t]{90}{\scriptsize \scalebox{1.2}{\textbf{BeNeRF}}}} &
		\includegraphics[width=1\textwidth,height=1in]{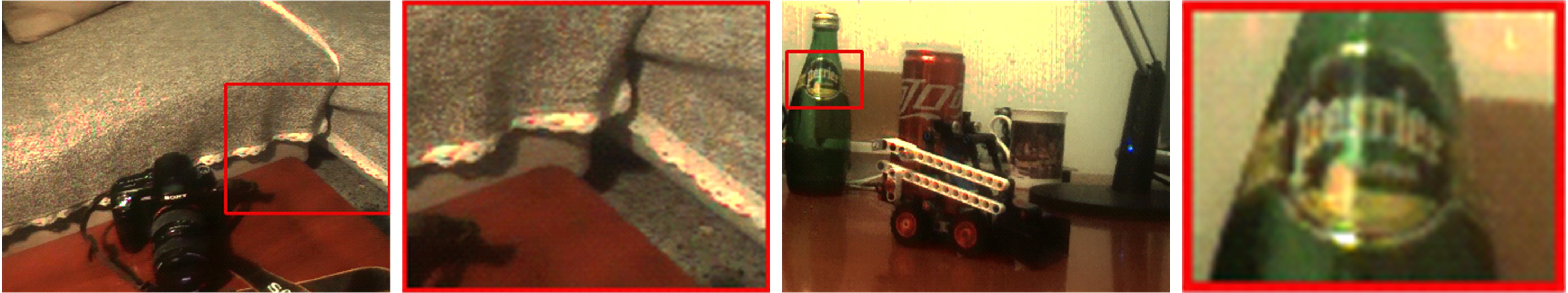}\\
		\specialrule{0em}{.02em}{.02em}
	\end{tabular}
	\caption{ {\textbf{Qualitative results of different methods with real-world datasets.}} Detailed qualitative comparison for “Camera” and "Lego" scene of real-world dataset.}
	\label{fig_e2nerf_real_1}
\end{figure*}

\begin{figure*}[t]
	\centering
	\begin{tabular}{cccccccc}
		\raisebox{0.31in}{\rotatebox[origin=t]{90}{\scriptsize \scalebox{1.2}{Input}}} &
		\includegraphics[width=1\textwidth,height=0.75in]{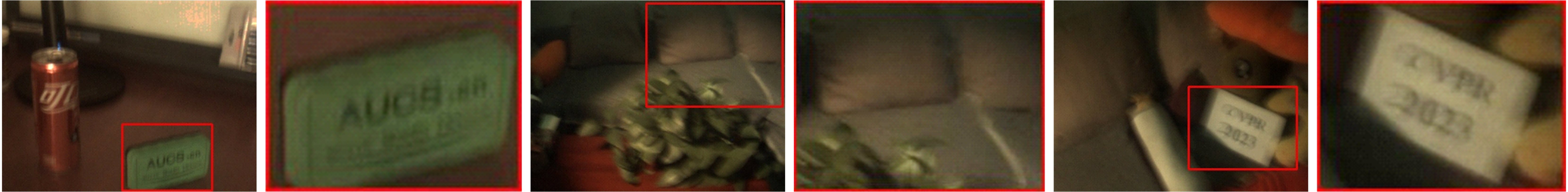}\\
		\specialrule{0em}{.02em}{.02em}
		\raisebox{0.31in}{\rotatebox[origin=t]{90}{\scriptsize \scalebox{1.2}{SRN} }} &
		\includegraphics[width=1\textwidth,height=0.75in]{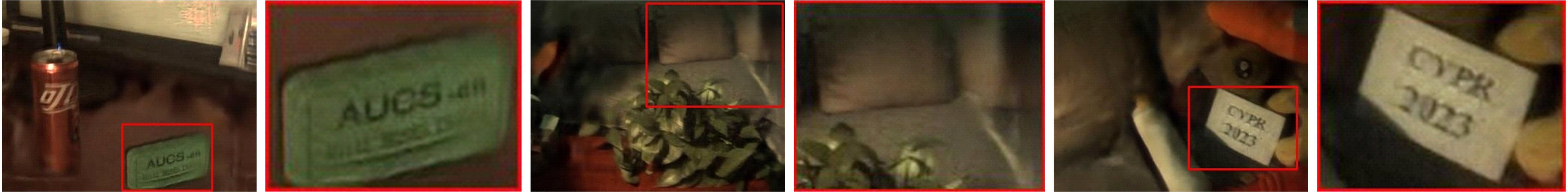}\\
		\specialrule{0em}{.02em}{.02em}
		\raisebox{0.31in}{\rotatebox[origin=t]{90}{\scriptsize \scalebox{1.2}{EDI}}} &
		\includegraphics[width=1\textwidth,height=0.75in]{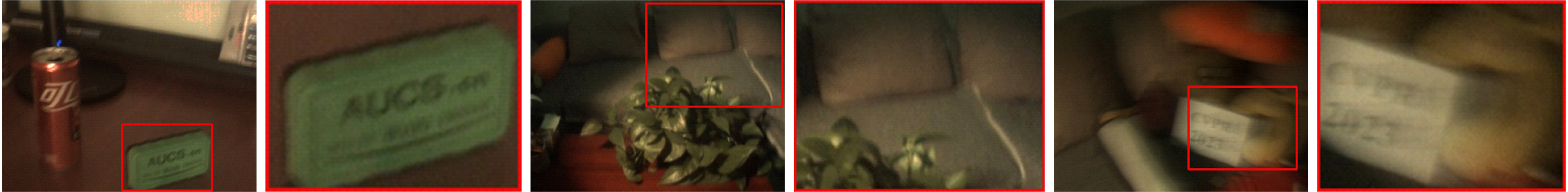}\\
		\specialrule{0em}{.02em}{.02em}
		\raisebox{0.31in}{\rotatebox[origin=t]{90}{\scriptsize \scalebox{1.2}{E$^2$NeRF}}} &
		\includegraphics[width=1\textwidth,height=0.75in]{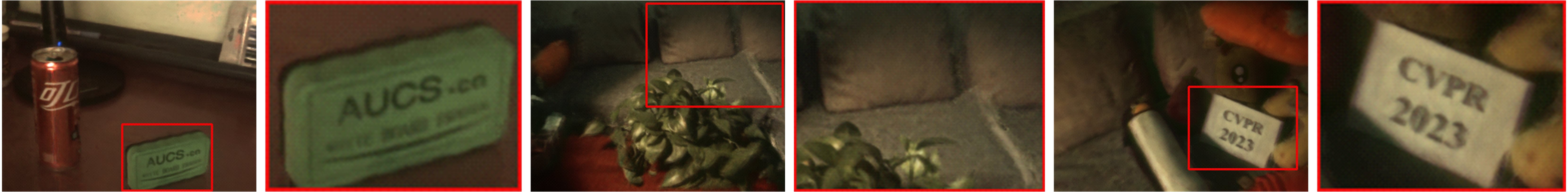}\\
		\specialrule{0em}{.02em}{.02em}
		\raisebox{0.31in}{\rotatebox[origin=t]{90}{\scriptsize \scalebox{1.2}{\textbf{BeNeRF}}}} &
		\includegraphics[width=1\textwidth,height=0.75in]{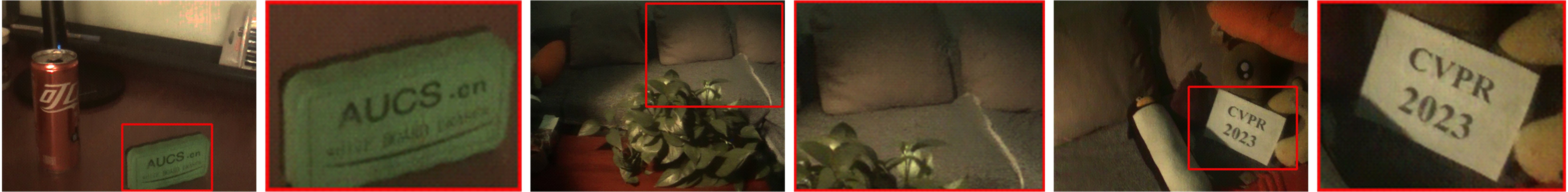}\\
		\specialrule{0em}{.02em}{.02em}
	\end{tabular}
	\caption{ {\textbf{Qualitative results of different methods with real-world datasets.}} Detailed qualitative comparison for “Letter”, "Plant" and "Toys" scene of real-world dataset.}
	\label{fig_e2nerf_real_2}
\end{figure*}

\figref{fig_e2nerf_real_1} and \figref{fig_e2nerf_real_2} illustrate comprehensive comparisons on real-world datasets. The results demonstrate the superior performance of our method on real datasets, attributed to its enhanced capability in modeling the physical process of motion-blurred imaging.

\begin{figure*}
	\centering
	\begin{tabular}{cccccccc}
		\raisebox{0.31in}{\rotatebox[origin=t]{90}{\scriptsize \scalebox{0.9}{Input}}} &
		\includegraphics[width=1\textwidth,height=0.7in]{fig/supplementary/e2nerf_synthetic_datasets/45-1.jpg}\\
		\specialrule{0em}{.02em}{.02em}
		\raisebox{0.31in}{\rotatebox[origin=t]{90}{\scriptsize \scalebox{0.9}{NeRF} }} &
		\includegraphics[width=1\textwidth,height=0.7in]{fig/supplementary/e2nerf_synthetic_datasets/45-2.jpg}\\
		\specialrule{0em}{.02em}{.02em}
		\raisebox{0.31in}{\rotatebox[origin=t]{90}{\scriptsize \scalebox{0.9}{E$^2$NeRF}}} &
		\includegraphics[width=1\textwidth,height=0.7in]{fig/supplementary/e2nerf_synthetic_datasets/45-4.jpg}\\
		\specialrule{0em}{.02em}{.02em}
		\raisebox{0.31in}{\rotatebox[origin=t]{90}{\scriptsize \scalebox{0.9}{\textbf{BeNeRF}}}} &
		\includegraphics[width=1\textwidth,height=0.7in]{fig/supplementary/e2nerf_synthetic_datasets/45-5.jpg}\\
		\specialrule{0em}{.02em}{.02em}
		\raisebox{0.31in}{\rotatebox[origin=t]{90}{\scriptsize \scalebox{0.9}{GT}}} &
		\includegraphics[width=1\textwidth,height=0.7in]{fig/supplementary/e2nerf_synthetic_datasets/45-6.jpg}\\
		\specialrule{0em}{.02em}{.02em}
	\end{tabular}
	\caption{ {\textbf{Qualitative results of different methods with synthetic datasets proposed by E$^2$NeRF.}} Detailed qualitative comparison for “Chair”, "Ficus" and "Hotdog" scene of synthetic dataset from E$^2$NeRF.}
	\label{fig_e2nerf_synthetic_1}
\end{figure*}

\section{Supplementary videos}

To showcase the effectiveness of our approach, we provide a supplementary video illustrating its capability to recover high-quality latent sharp video from a single blurry image and corresponding event stream, which encapsulates rich temporal information. These videos are available on the \href{https://akawincent.github.io/BeNeRF}{project page}. Furthermore, our results highlight the superior performance of our method in comparison to previous state-of-the-art approaches.

\clearpage
%
%
\bibliographystyle{splncs04}
\bibliography{reference}
\end{document}